\documentclass[10pt,twocolumn,letterpaper]{article}

\usepackage[pagenumbers]{iccv} 

\definecolor{iccvblue}{rgb}{0.21,0.49,0.74}
\usepackage[pagebackref,breaklinks,colorlinks,allcolors=iccvblue]{hyperref}

\def\paperID{8067} 
\def\confName{ICCV}
\def\confYear{2025}

\usepackage{amsmath,amssymb}
\usepackage{multirow}
\usepackage{arydshln}
\usepackage{units}
\usepackage{bbm}
\usepackage{xcolor}

\title{CoMatch: Dynamic Covisibility-Aware Transformer for Bilateral Subpixel-Level Semi-Dense Image Matching}

\author{
	Zizhuo Li \quad
	Yifan Lu \quad
	Linfeng Tang \quad
	Shihua Zhang \quad
	Jiayi Ma\footnotemark[1] \footnotetext[1]{\thanks{Corresponding authors.}}\\
	Wuhan University, China \quad \\
	{\tt\small  \{zizhuo\_li, lyf048\}@whu.edu.cn, \{linfeng0419,  suhzhang001, jyma2010\}@gmail.com}
}

\begin{document}
\maketitle
\begin{abstract}
	This prospective study proposes CoMatch, a novel semi-dense image matcher with dynamic covisibility awareness and bilateral subpixel accuracy. Firstly, observing that modeling context interaction over the entire coarse feature map elicits highly redundant computation due to the neighboring representation similarity of tokens, a covisibility-guided token condenser is introduced to adaptively aggregate tokens in light of their covisibility scores that are dynamically estimated, thereby ensuring computational efficiency while improving the representational
	capacity of aggregated tokens simultaneously. Secondly, considering that feature interaction with massive non-covisible areas is distracting, which may degrade feature distinctiveness, a covisibility-assisted attention mechanism is deployed to selectively suppress irrelevant message broadcast from non-covisible reduced tokens, resulting in robust and compact attention to relevant rather than all ones. Thirdly, we find that at the fine-level stage, current methods adjust only the target view's keypoints to subpixel level, while those in the source view remain restricted at the coarse level and thus not informative enough, detrimental to keypoint location-sensitive usages. A simple yet potent fine correlation module is developed to refine the matching candidates in both source and target views to subpixel level, attaining attractive performance improvement. Thorough experimentation across an array of public benchmarks affirms CoMatch’s promising accuracy, efficiency, and generalizability.
\end{abstract}
\section{Introduction}
\label{sec:intro}
\begin{figure}[t]
	\centering
	\includegraphics[width=0.82\linewidth]{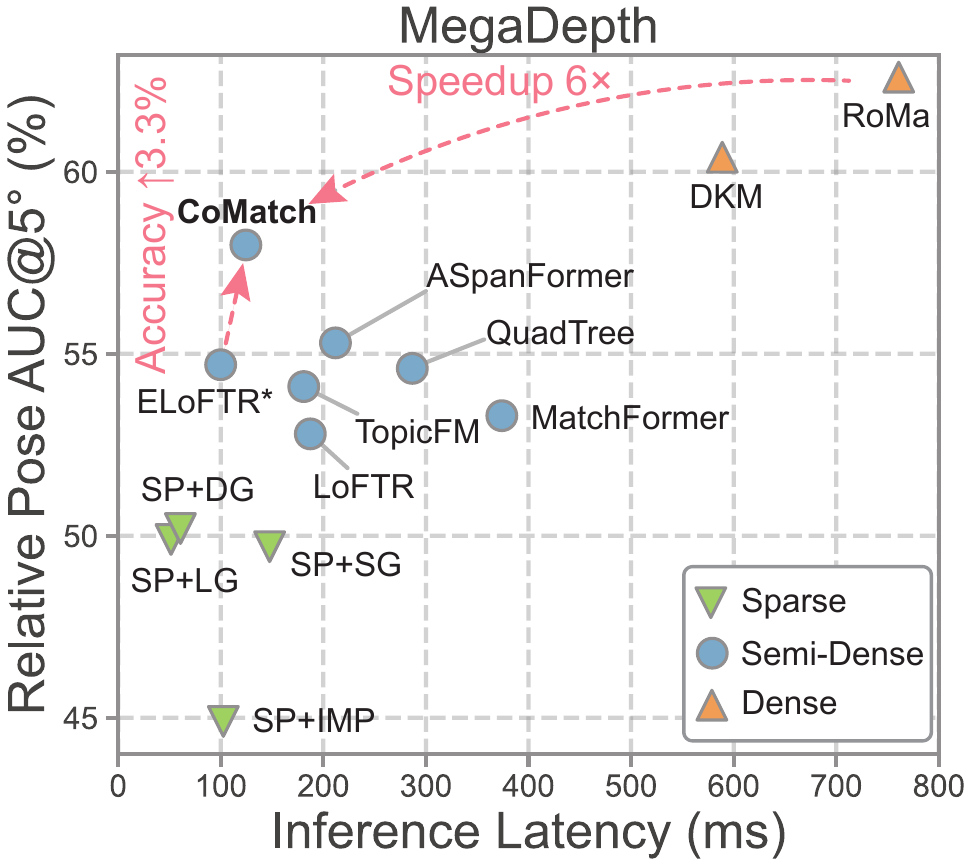}
	\vspace{-0.08in}
	\caption{\textbf{Image matching accuracy and efficiency on MegaDepth.} CoMatch achieves remarkably better accuracy than both sparse (\raisebox{-0.5ex}{\includegraphics[height=1.em]{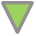}}) and semi-dense (\raisebox{-0.4ex}{\includegraphics[height=1.em]{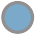}}) matchers with a commendable speed. Compared with dense matcher RoMa (\raisebox{-0.25ex}{\includegraphics[height=1.em]{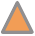}}), our method is $\sim6\times$ faster with comparable performance.}\label{performance}
\end{figure}

Image matching aims to establish reliable correspondences between a pair of images depicting the same visual content, holding a fundamentally special place for numerous 3D visual tasks, such as Simultaneous Localization and Mapping (SLAM)~\cite{mur2015orb,mur2017orb}, Structure-from-Motion (SfM)~\cite{schonberger2016structure,he2024detector}, and novel view synthesis~\cite{mildenhall2021nerf,kerbl20233d}. Conventionally, the pipeline starts with detecting keypoints~\cite{rosten2006machine} and constructing their associated visual descriptors~\cite{detone2018superpoint}, so that point-to-point matches can be generated by heuristic tricks~\cite{lowe2004distinctive} or learning-based matchers~\cite{sarlin2020superglue,lindenberger2023lightglue}. Albeit efficient, detector-based methods still struggle to robustly detect repeatable keypoints and effectively eliminate the visual ambiguity of descriptors across sophisticated scenarios, like poor textures, repetitive patterns, and large displacements.

To cope with this limitation, a line of research~\cite{sun2021loftr,jiang2021cotr,tangquadtree,wang2022matchformer,wang2024efficient,edstedt2023dkm,edstedt2024roma} resorts to Transformer backbone~\cite{vaswani2017attention,dosovitskiy2020image} to explore the feasibility of directly matching local features on dense grids rather than sparse locations (\emph{i.e.}, a detector-free matching paradigm), so that more plentiful context can be leveraged while the keypoint detection stage can be eschewed spontaneously. Pioneeringly, LoFTR~\cite{sun2021loftr} initially aggregates global intra- and inter-image context at the coarse level using self- and cross-attention, and then crops fixed-size feature patches based on coarse matches for fine-level refinement, achieving encouraging performance. Built on the success of such coarse-to-fine matching paradigm, numerous LoFTR’s follow-ups~\cite{chen2022aspanformer,wang2022matchformer,giang2023topicfm} have emerged and made notable contributions to further improve the matching accuracy. However, they struggle with limited efficiency due to the intensive computation at the coarse level. Most recently, ELoFTR~\cite{wang2024efficient} develops an aggregated attention mechanism by conducting message passing across reduced tokens, thus introducing considerable efficiency gain. Yet, it delivers comparable matching accuracy, primarily due to
1) the compromised representational capacity of reduced tokens and 2) context aggregation without considering the differences in reliability among reduced tokens, involving an extensive range of non-covisible regions that are not conducive to learning discriminative features. Furthermore, such detector-free methods suffer from a prevalent shortcoming: at the fine-level phase, only keypoints in the target view are refined to subpixel level while those in the source view are spatially restricted to pixel-level accuracy, severely hindering the matching performance and particularly unfriendly to keypoint location-sensitive usages. Overall, long lines of previous research are either insufficiently accurate or unacceptably efficient.

\begin{figure}[t]
	\centering
	\includegraphics[width=\linewidth]{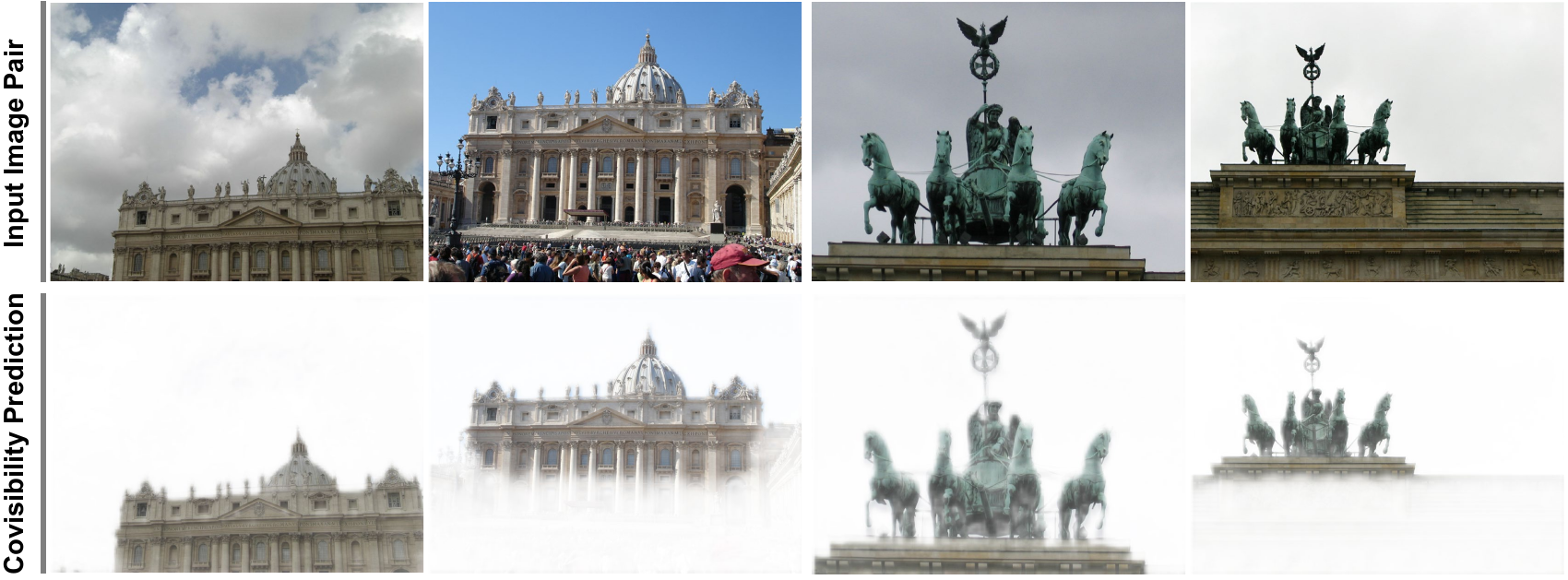}
	\caption{\textbf{Visualization of covisibility prediction.}	We first bilinearly up-sample the covisibility score map to match the original image resolution, and then multiply it with the input image.}\label{covisibility}
\end{figure}

To remedy the above issues, this paper proposes CoMatch, a detector-free matcher that achieves the best of both worlds in accuracy and efficiency, as shown in Fig.~\ref{performance}. Our key innovations lie in designing a dynamic covisibility-aware Transformer for efficient and robust coarse-level feature transformation and a bilateral subpixel regression module for fine-level correspondence location refinement. More concretely, based on the pivotal observations that 1) intensively conducting message passing across the entire coarse feature map is unnecessary as neighboring tokens share similar representations and 2) tokens in covisible areas embrace fruitful geometric and visual cues, a covisibility-guided token condenser is presented, which dynamically predicts the covisibility of each token (as visualized in Fig.~\ref{covisibility}) and then guides adaptive token selection accordingly. This technique not only optimizes computational efficiency but also empowers aggregated tokens with powerful representational capability. On top of that, we find that accepting context from non-covisible regions is meaningless and distracting, which poses a substantial barrier to distinctive feature learning. In this regard, an ideal attention mechanism should adaptively suppress noisy message interaction with non-covisible aggregated tokens. Following this train of thought, we devise a covisibility-assited attention mechanism, which helps our network to sharply focus on covisible regions while discarding irrelevant ones, allowing for robust and compact covisibility-dependent context aggregation. Together, these above two components constitute our dynamic covisibility-aware Transformer. Besides, we notice that at the matching refinement stage, existing methods only achieve unilateral subpixel accuracy, adverse to keypoint location-sensitive usages. To handle this downside, we directly optimize a task-oriented epipolar loss to simultaneously refine both views' keypoints to subpixel level in a symmetric manner.  This simplistic design yields geometrically consistent matches, attaining surprising performance.

In summary, this paper has the following contributions:
\begin{itemize}
	\item We propose a dynamic covisibility-aware Transformer for coarse-level feature transformation. Thanks to elaborate token condensing and attention modules, it performs context interaction efficiently, robustly, and compactly.
	\item We propose a novel correlation module for fine-level match refinement. With epipolar geometry as supervision, it achieves bilateral subpixel accuracy for two camera views, benefiting keypoint location-sensitive usages.
	\item Extensive experiments showcase that our method not only reaches state-of-the-art scores on multiple visual tasks, beating long lines of prior work, but also maintains exceptional efficiency, conducive to real-world applications.
\end{itemize}

\section{Related Work}
\label{sec:related_work}
\subsection{Detector-Based Image Matching}
Detector-based image matching has been studied since the early years and showcases unbroken popularity. Traditional methods leverage hand-engineered detector-descriptors~\cite{lowe2004distinctive, rublee2011orb,bay2008speeded} to produce 2D keypoints associated with visual descriptions per image and then matching them for correspondence generation with nearest neighbor search (NN) and its variants, \emph{e.g.}, mutual nearest neighbor check (MNN) and Lowe’s ratio test (RT)~\cite{lowe2004distinctive}. Recently, integrating deep learning
techniques into individual stages of such a pipeline has received inspiring success. Particularly, one thread of studies~\cite{detone2018superpoint,dusmanu2019d2,revaud2019r2d2,luo2020aslfeat,wang2023attention,zhao2023aliked,xue2023sfd2} focuses on improving the robustness of local features under illumination variation and viewpoint change. Contemporarily, the other thread~\cite{sarlin2020superglue, shi2022clustergnn,xue2023imp,jiang2023improving,deng2024resmatch, zhang2024diffglue} specializes in learning to match local features, mitigating the limitation of vanilla NN matches. Groundbreakingly, SuperGlue~\cite{sarlin2020superglue} takes two groups of keypoints
and their corresponding local descriptors as input and outputs learned correspondences using an attentional graph neural network~\cite{wu2020comprehensive} with densely connected intra- and inter-image graphs. However, its impressive performance comes with high computational complexity of $\mathcal{O}(N^2)$, where $N$ is the keypoint number, profoundly restricting its applicability to latency-sensitive tasks. Numerours innovations, like SGMNet~\cite{chen2021learning} and LightGlue~\cite{lindenberger2023lightglue}, have endeavored to remedy this issue by leveraging a set of reliable matches as message bottleneck and adjusting the network's depth and width based on the image pair's difficulty, respectively, thus achieving higher efficiency and comparable accuracy. Notably, the upper bound on the performance of these learnable matchers is limited by the quality of detected keypoints. But unfortunately, robust detection of repeatable keypoints still encounters formidable challenges, especially in the case of poor textures.

\subsection{Detector-Free Image Matching}
Detector-free image matching~\cite{sun2021loftr,huang2023adaptive,cai2024prism,chenecomatcher} enjoys an end-to-end pipeline that directly searches matches from the raw image pair without an explicit keypoint detection phase, broadly categorized into semi-dense and dense matching.

Earlier semi-dense matching methods are represented by NCNet~\cite{rocco2018neighbourhood} that represents all underlying matches as a 4D cost volume, where 4D convolutions are used for regularization and imposing neighborhood consensus on all matches. Encouraged by NCNet, follow-ups innovate on 4D cost volume construction and calculation~\cite{rocco2020efficient,li2020dual}. Recently, LoFTR~\cite{sun2021loftr} harnesses Transformer~\cite{vaswani2017attention} to model long-range dependencies, where linear self- and cross-attention blocks~\cite{katharopoulos2020transformers} are used to update cross-view features with manageable computational burden. Built on the success of LoFTR, a large bunch of variants have emerged and shown greater matching accuracy or efficiency. For instance, QuadTree~\cite{tangquadtree} constructs token pyramids and performs context interaction in a coarse-to-fine manner. AspanFormer~\cite{chen2022aspanformer} develops a global-and-local attention strategy to conduct multi-scale message passing and integrates flow estimation to guide local attention spanning.  AffineFormer~\cite{chen2024affine} improves upon AspanFormer by regularizing the regressed intermediate flows with affine transformation. TopicFM~\cite{giang2023topicfm} presents the perspective of latent topic modeling, where semantically similar features are assigned to the identical topic and then messages are passed within each topic. Despite the performance improvement concerning accuracy or efficiency offered by these semi-dense matching methods, it remains an open question of how to strike a satisfying balance between these two key elements.

Dense matching methods~\cite{truong2020glu,jiang2021cotr,truong2021learning,truong2023pdc,ni2023pats,edstedt2023dkm,edstedt2024roma} regress a dense flow field relating two-view images, coupled with a pixel-wise confidence map indicating the reliability and accuracy of the estimation, bridging the task of local feature matching and optical flow. However, they are generally much slower than sparse and semi-dense methods, and it is currently infeasible to extend them into high-resolution cases due to their computationally intensive nature.

\section{Methodology}
\subsection{Overview}
\begin{figure*}[t]
	\centering
	\includegraphics[width=.99\linewidth]{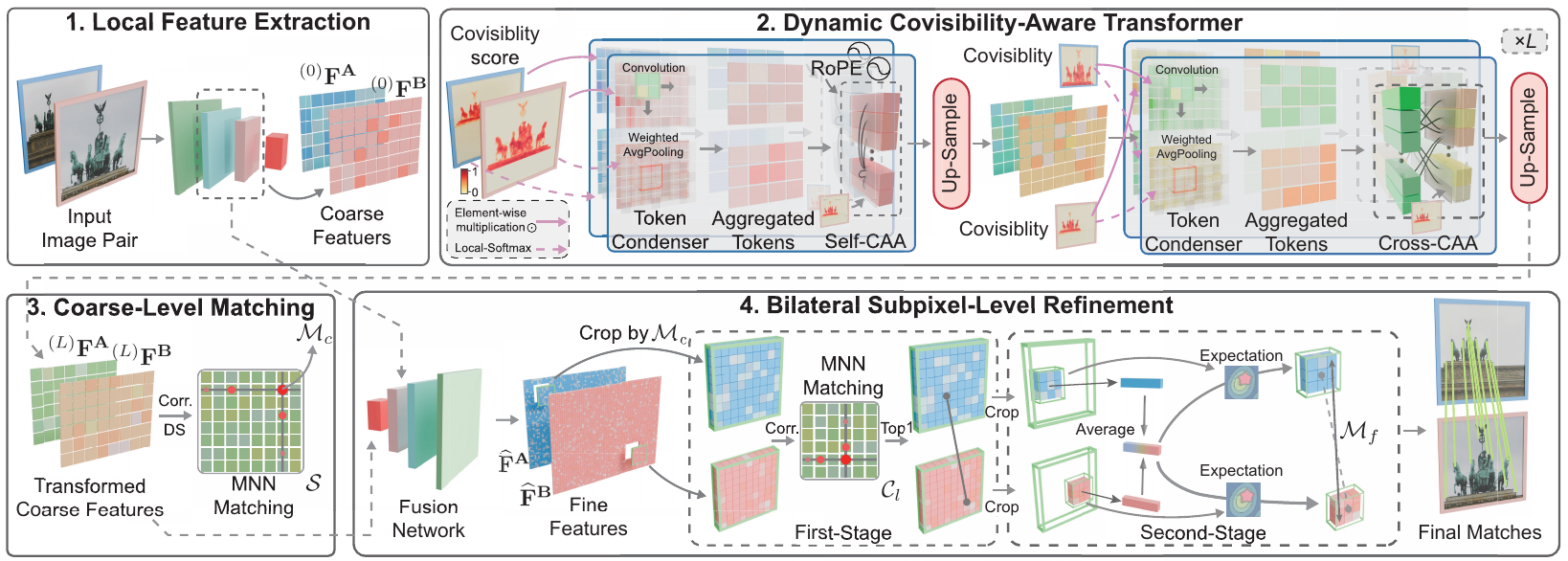}
	\vspace{-0.05in}
	\caption{\textbf{Pipeline overview.} \textbf{(1)} Given a pair of images, a CNN network extracts coarse features $^{(0)}\mathbf{F}^{\mathbf{A}}$ and $^{(0)}\mathbf{F}^{\mathbf{B}}$, alongside fine ones. \textbf{(2)} Dynamic covisibility-aware Transformer is stacked $L$ times to conduct efficient, robust, and compact context interaction for coarse feature transformation. \textbf{(3)} Transformed coarse features are correlated, followed by a dual-softmax (DS) operation to yield the assignment matrix $\mathcal{S}$, where mutual nearest neighbor (MNN) matching is used to establish coarse matches $\mathcal{M}_c$. \textbf{(4)} Fine distinctive features $\widehat{\mathbf{F}}^\mathbf{B}$ and $\widehat{\mathbf{F}}^\mathbf{B}$ at the original resolution are derived by progressively fusing $^{(L)}\mathbf{F}^{\mathbf{A}}$ and $^{(L)}\mathbf{F}^{\mathbf{B}}$ with backbone features at $\nicefrac{1}{4}$ and $\nicefrac{1}{2}$ resolutions. Later, feature patches centered on $\mathcal{M}_c$ are cropped, followed by a two-stage refinement to produce fine matches $\mathcal{M}_f$ with bilateral subpixel accuracy.}\label{overview}
\end{figure*}

Suppose we are given a pair of images $\mathbf{A}$ and $\mathbf{B}$ depicting the
same 3D structure, our primary objective is to establish a set of superior-quality correspondences between them. For this purpose, we resort to a coarse-to-fine semi-dense matching pipeline, which initially identifies coarse-level correspondences on down-sampled feature maps and then refines them to subpixel positions for high matching accuracy. The overview of CoMatch is presented in Fig.~\ref{overview}.

\subsection{Local Feature Extraction}
Instead of adopting a computationally intensive ResNet-FPN~\cite{lin2017feature} commonly used in detector-free matchers, we discarding its up-sampling components, retaining only its down-sampling ones to extract intermediate down-sampled coarse features $^{(0)}\mathbf{F}^{\mathbf{A}}$, $^{(0)}\mathbf{F}^{\mathbf{B}}$ at $\nicefrac{1}{8}$ image resolution,  along with fine features at $\nicefrac{1}{4}$ and $\nicefrac{1}{2}$ image resolutions for later coarse-to-fine matching, as shown in Fig.~\ref{overview} (1). This way accelerates the inference process while maintaining model performance utmostly (\emph{cf}. Sec.~\ref{ablation_studies}).

\subsection{Dynamic Covisibility-Aware Transformer}
After feature extraction, we apply  the dynamic covisibility-aware Transformer (DCAT) $L$ times for local feature transformation. This process enhances the feature distinctiveness, facilitating easier matching. We denote the transformed coarse-level features as $^{(L)}\mathbf{F}^{\mathbf{A}}$, $^{(L)}\mathbf{F}^{\mathbf{B}}$.

\noindent{\textbf{Preliminaries.}} Prior to delving into DCAT, we first present a succinct overview of the commonly utilized vanilla attention~\cite{vaswani2017attention}. Concretely, vanilla attention takes a set of query ($\mathbf{Q}$), key ($\mathbf{K}$), and value ($\mathbf{V}$) vectors as input and outputs a weighted sum of $\mathbf{V}$, according to the weighting matrix derived from the similarity matrix between $\mathbf{Q}$ and $\mathbf{K}$. Note that $\mathbf{Q}$ is linear projection of $\rm{Img2Seq}(\mathbf{F}^\mathbf{I})$, and $\mathbf{K}$, $\mathbf{V}$ are linear projections of $\rm{Img2Seq}(\mathbf{F}^\mathbf{J})$, where $\rm{Img2Seq}(\cdot)$ flattens a feature map into a feature sequence and $\mathbf{I},\mathbf{J}\in\{\mathbf{A}, \mathbf{B}\}$. Mathematically, vanilla attention is defined as: $\rm{VAtt}(\mathbf{Q}, \mathbf{K},\mathbf{V})=\rm{Softmax}(\mathbf{Q}\mathbf{K}^{\top})\mathbf{V}$. However, directly applying vanilla attention to dense local features is impractical, due to the quadratic increase in the size of the weighting matrix $\rm{Softmax}(\mathbf{Q}\mathbf{K}^{\top})$ \emph{w.r.t.} the image size, causing an excessive computational burden. To resolve this issue, prior research adopts linear attention~\cite{katharopoulos2020transformers} for message passing, significantly lowering the computational complexity from quadratic to linear. Unfortunately, its manageable computational overhead comes at the cost of compromising the representational ability~\cite{cai2022efficientvit} and matching accuracy~\cite{tangquadtree}. Thus, we introduce DCAT, taking advantage of each token's covisibility for efficient and robust message passing.

\noindent{\textbf{Covisibility-Guided Token Condenser.}} Previous methods typically perform context interaction across the entire coarse feature map, which brings in heavily redundant computation. This is because 1) neighboring query tokens hold similar attention information and 2) each query token centralizes the bulk of attention weights on a small number of key tokens. That is to say, we can aggregate neighboring tokens to avert redundant attention computation. A simplistic manner to achieve this is to leverage a depth-wise convolution network for $\mathbf{F}^\mathbf{I}$ and a max pooling layer for $\mathbf{F}^\mathbf{J}$~\cite{wang2024efficient}:
\begin{equation}\label{aggregated_attention}
	\widetilde{\mathbf{F}}^\mathbf{I}=\rm{Conv2D}(\mathbf{F}^\mathbf{I}),\; \widetilde{\mathbf{F}}^\mathbf{J}=\rm{MaxPool}(\mathbf{F}^\mathbf{J}),
\end{equation}
where $\rm{Conv2D}$ is a strided depth-wise convolution with a kernel size of $s\times s$, congruent with that of the max pooling layer, and $\mathbf{I},\mathbf{J}\in\{\mathbf{A}, \mathbf{B}\}$. By doing so, the number of tokens in $\widetilde{\mathbf{F}}^\mathbf{I}$ and $\widetilde{\mathbf{F}}^\mathbf{J}$ is reduced by $s^2$, remarkably promoting the attention efficiency. However, as mentioned in Sec.~\ref{sec:intro}, the representational capacity of reduced tokens is sacrificed substantially, causing only comparable matching accuracy.

To eradicate this defect, considering that tokens within covisible regions inherently contain richer geometric and visual cues than those in non-covisible regions, we propose a covisibility-guided token condenser (CGTC) that adaptively condenses tokens on the basis of their covisibility scores dynamically yielded by a lightweight classifier, \emph{i.e.}, a multi-layer perceptron (MLP). At the start of the $\ell$-th DCAT block, we compute, for each token, a covisibility score as\footnote{Notably, for the first DCAT block, predicting the covisibility score for each token based on coarse-level features $^{(0)}\mathbf{F}^{\mathbf{A}}$ and $^{(0)}\mathbf{F}^{\mathbf{B}}$ is infeasible, as they only encode intra-image geometric and visual cues without perceiving inter-image ones. Therefore, we assume $^{(1)}\mathbf{C}^{\mathbf{A}},^{(1)}\mathbf{C}^{\mathbf{B}}=\mathbf{1}$.}:
\begin{equation}\label{covisibility_prediction}
	^{(\ell)}\mathbf{C}_i=\rm{Sigmoid}(\rm{MLP}(^{(\ell-1)}\mathbf{F}_{\mathit{i}}))\in[0,1].
\end{equation}
This score encodes the likelihood of token $i$ to be in covisible regions, as illustrated in Fig.~\ref{covisibility}. A token that is not observed in the other image, \emph{e.g.}, when occluded, is non-covisible and thus has $\mathbf{C}_i\rightarrow 0$.

After that, for $^{(\ell-1)}\mathbf{F}^\mathbf{I}$, we propose to recalibrate spatial-wise responses using the predicted covisibility scores, followed by a depth-wise convolution for token condensing, thus emphasizing informative covisible tokens while suppressing less useful non-covisible ones:
\begin{equation}\label{condensing_query}
	^{(\ell-1)}\widetilde{\mathbf{F}}^\mathbf{I}=\rm{Conv2D}(^{(\ell-1)}\mathbf{F}^\mathbf{I}\odot^{(\ell)}\mathbf{C}^\mathbf{I}), \;\mathbf{I}\in\{\mathbf{A}, \mathbf{B}\},
\end{equation}
where $\odot$ denotes element-wise multiplication.

For $^{(\ell-1)}\mathbf{F}^\mathbf{J}$, simply utilizing max pooling for token condensing inevitably entails the loss of considerable valuable local information. To obviate this, one may consider average pooling as an alternative. However, such a practice treats each token in a neighborhood equally, which is problematic, as it overlooks the adverse impact of non-covisible tokens, rendering the representational capacity of condensed tokens not powerful. Ideally, the contextual information embedded in these condensed tokens should exclusively originate from covisible ones. To approach this target, we propose to weigh the importance of each token in a local region according to their covisibility scores, ensuring purer covisible context embedding. For this purpose, we generalize average pooling over a specific $s\times s$ local region $\mathcal{N}_s$ to a weighted formulation as follows:
\begin{equation}\label{condensing_key_value}
	\mathcal{G}(^{(\ell-1)}\mathbf{F}^\mathbf{J},^{(\ell)}\mathbf{C}^{\mathbf{J}},\mathcal{N}_s)=\sum\nolimits_{j\in\mathcal{N}_\mathit{s}}\rm{Softmax}(^{(\ell)}_{\mathcal{N}_\mathit{s}}\mathbf{C}^{\mathbf{J}})_\mathit{j},
\end{equation}
where $\mathcal{G}(\cdot, \cdot, \cdot)$ denotes the weighted average pooling, $^{(\ell)}_{\mathcal{N}_\mathit{s}}\mathbf{C}^{\mathbf{J}}=\{^{(\ell)}\mathbf{C}_{\mathit{k}}^{\mathbf{J}}\}_{k\in\mathcal{N}_\mathit{s}}$, and $\mathbf{J}\in\{\mathbf{A}, \mathbf{B}\}$.

In this way, we can achieve the best of both worlds in the computational efficiency of subsequent attention and the representational capacity of condensed tokens.

\noindent{\textbf{Covisibility-Assisted Attention.}}
Thanks to the significantly reduced token size achieved by the CGTC module, we can alternate to vanilla attention that offers greater representational power than linear attention, for inter- and intra-image context interaction. However, as analyzed in Sec.~\ref{sec:intro}, in the context of local feature matching, numerous tokens are located in non-covisible regions, due to factors like occlusion and viewpoint changes, and thus irrelevant for message passing. Retrieving information from them is unnecessary and distracting, ultimately impairing the feature discriminability. In this regard, we present a covisibility-assisted attention (CAA) module, as shown in Fig.~\ref{overview}~(2). Concretely, at layer $\ell$, we first align the size of covisibility scores with reduced tokens before using them to assist attention. Crucially, if any token in a local region is covisible, the corresponding aggregated token should be considered covisible. Thus, max-pooling is adopted to achieve this:
\begin{equation}\label{covisibility_aggregation}
	^{(\ell)}\widetilde{\mathbf{C}}^\mathbf{J}=\rm{MaxPool}(^{(\ell)}\mathbf{C}^\mathbf{J}),\;\mathbf{J}\in\{\mathbf{A}, \mathbf{B}\}.
\end{equation}
Then, a weighted version of vanilla attention is introduced as follows, for covisibility-dependent context interaction:
\begin{equation}
	\begin{aligned}\label{covisibility_attention}
		\rm{WAtt}(\mathbf{Q}, \mathbf{K}, \mathbf{V},^{(\ell)}&\widetilde{\mathbf{C}}^\mathbf{J}) = \rm{Softmax}(\mathbf{Q} \mathbf{K}^{\top}) \mathbf{W} \mathbf{V}, \\
		\mathbf{W} =\ & \rm{Diag}(\rm{Img2Seq}(^{(\ell)}\widetilde{\mathbf{C}}^\mathbf{J})).
	\end{aligned}
\end{equation}
Ultimately, the transformed feature map is up-sampled and fused with $^{(\ell-1)}\mathbf{F}^\mathbf{I}$ to yield the final feature map.

Importantly, with the help of predicted covisibility scores, the proposed CAA module is capable of adaptively suppressing noisy information flow from non-covisible condensed tokens and sharply concentrating on covisible ones, facilitating robust and compact message passing.

\noindent{\textbf{Positional Encoding.}} We adopt RoPE~\cite{su2024roformer} for positional encoding, which allows for retrieving point $j$ located at a learned relative position from $i$. Notably, we just apply RoPE to self-attention, since incorporating any relative positional information across images is meaningless.

\subsection{Coarse-Level Match Determination}
After being updated by $L$ DCAT blocks, we can obtain the transformed coarse-level feature maps $^{(L)}\mathbf{F}^{\mathbf{A}}$ and $^{(L)}\mathbf{F}^{\mathbf{B}}$, which are used to construct the correlation matrix $\mathcal{C}$ by $\mathcal{C}(i,j)=\tau\langle^{(L)}\mathbf{F}^{\mathbf{A}}_{\mathit{i}}, ^{(L)}\mathbf{F}^{\mathbf{B}}_{\mathit{j}}\rangle$, where $\tau$ is a learnable coefficient and $\langle\cdot,\cdot\rangle$ denotes inner product, followed by a dual-softmax operation to generate the assignment matrix $\mathcal{S}$~\cite{sun2021loftr}. Lastly, we derive the coarse-level matches $\mathcal{M}_c$ as follows: $\mathcal{M}_c=\{(i,j)\;\vert\;\forall(i,j)\in\rm{MNN}(\mathcal{S}), \mathcal{S}(\mathit{i},\;\mathit{j})\geq\theta_c)\}$,
where $\theta_c$ is used to prune unreliable coarse-level matches.


\subsection{Bilateral Subpixel-Level Refinement}
Following the coarse-level matching candidates $\mathcal{M}_c$, we additionally deploy a bilateral subpixel-level refinement (BSR) module to further refine more detailed matching results, \emph{i.e.}, subpixel accuracy, for both source and target images. It is fleshed out by a progressive feature fusion module and a two-stage feature correlation layer for the final fine-level matches $\mathcal{M}_f$, as overviewed in Fig.~\ref{overview} (4).

\noindent{\textbf{Progressive Feature Fusion.}} This module aims to provide cross-view perceived fine features for upcoming refinement. It progressively fuses $^{(L)}\mathbf{F}^{\mathbf{A}}$ and $^{(L)}\mathbf{F}^{\mathbf{B}}$ with previously extracted backbone feature maps at $\nicefrac{1}{4}$ and $\nicefrac{1}{2}$ image resolutions through convolution and up-sampling,  yielding fine-level $\widehat{\mathbf{F}}^\mathbf{A}$ and $\widehat{\mathbf{F}}^\mathbf{B}$ at the full resolution. Later, two sets of fine-level feature patches centered on coarse-level matches $\mathcal{M}_c$ are cropped from $\widehat{\mathbf{F}}^\mathbf{A}$ and $\widehat{\mathbf{F}}^\mathbf{B}$, followed by a vanilla cross attention to further consolidate their discriminability.

\noindent{\textbf{Two-Stage Correlation.}}
We adopt a two-stage correlation paradigm for match refinement. At the first stage, following~\cite{wang2024efficient}, for every coarse-level match in $\mathcal{M}_c$, we first densely correlate its corresponding two fine-level feature patches to derive a local patch correlation matrix $\mathcal{C}_l$, on which we then utilize an MNN strategy to yield intermediate pixel-level fine matches and retain only the one with the highest correlation score to restrict the total match number.

The second-stage refinement phase adopts a correlation-then-expectation manner for further refinement. Unlike LoFTR~\cite{sun2021loftr} and many other detector-free matchers~\cite{wang2022matchformer,chen2022aspanformer,chenecomatcher} that only reach unilateral subpixel accuracy by fixing the keypoints in the source image to pixel level and solely refining keypoints in the target image to subpixel level, we aim to refine these pixel-level matching candidates in both source and target views to subpixel accuracy in a symmetric way. Specifically, as the matching accuracy has been substantially enhanced, enabling the utilization of a tiny local window for correlation and expectation with a maximum suppression, hence for each pixel-level match $(\widehat{i},\widehat{j})$, we first crop $3\times3$ local feature patches $\widehat{\mathbf{P}}^\mathbf{A}_{\widehat{i}}$ and $\widehat{\mathbf{P}}^\mathbf{B}_{\widehat{j}}$ centered on $\widehat{i}$ and $\widehat{j}$ from $\widehat{\mathbf{F}}^\mathbf{A}$ and $\widehat{\mathbf{F}}^\mathbf{B}$, respectively. Then, we couple its corresponding features $\widehat{\mathbf{F}}^\mathbf{A}_{\widehat{i}}$ and $\widehat{\mathbf{F}}^\mathbf{B}_{\widehat{j}}$ to generate a match representation vector $\mathbf{f}_{(\widehat{i},\widehat{j})}$ as $\mathbf{f}_{(\widehat{i},\widehat{j})}=\nicefrac{(\widehat{\mathbf{F}}^\mathbf{A}_{\widehat{i}}+\widehat{\mathbf{F}}^\mathbf{B}_{\widehat{j}})}{2}$. Finally, we concurrently correlate $\mathbf{f}_{(\widehat{i},\widehat{j})}$ with $\widehat{\mathbf{P}}^\mathbf{A}_{\widehat{i}}$ and $\widehat{\mathbf{P}}^\mathbf{B}_{\widehat{j}}$, followed by a soft-argmax operation to  regress the residual matching flows for both views, which adjust $(\widehat{i},\widehat{j})$ to $(\widehat{i}',\widehat{j}')$, respectively. By doing so,
the final fine-level matches $\mathcal{M}_f=\{(\widehat{i}', \widehat{j}')\}$ are endowed with bilateral subpixel accuracy, friendly to keypoint location-sensitive usages.

\subsection{Supervision}
To supervise CoMatch, we adopt a joint loss $\mathcal{L}$, consisting of 1) coarse-level matching loss $\mathcal{L}_c$, 2) fine-level refinement loss $\mathcal{L}_f$, and 3) covisibility estimation loss $\mathcal{L}_{covi}$.

\noindent{\textbf{Coarse-Level Matching Supervision.}} In line with~\cite{sun2021loftr,chen2022aspanformer}, we establish ground-truth coarse matches $\mathcal{M}_c^{gt}$ by warping grid-level points from $\mathbf{A}$ to $\mathbf{B}$ using camera poses and depth maps, and supervise the assignment matrix $\mathcal{S}$ by minimizing the negative log-likelihood loss over locations in $\mathcal{M}_c^{gt}$:
\begin{equation}\label{coarse_loss}
	\mathcal{L}_c=-\frac{1}{\vert\mathcal{M}_c^{gt}\vert}\sum\nolimits_{(i,j)\in\mathcal{M}_c^{gt}}\log(\mathcal{S}(i,j)).
\end{equation}

\noindent{\textbf{Fine-Level Refinement Supervision.}} We supervise two phases of the fine-level refinement module separately. Following~\cite{wang2024efficient}, the first-stage loss $\mathcal{L}_{f1}$ minimizes the negative log-likelihood loss over each local patch correlation matrix $\mathcal{C}_l$ grounded on pixel-level ground-truth matches, analogous to the coarse-level loss. To ensure final bilateral subpixel-level matches $\mathcal{M}_f$ with geometric consistency, we supervise the second stage with an epipolar geometry loss $\mathcal{L}_{f2}$:
\begin{align}\label{coarse_loss}
	\mathcal{L}_{f2}=&\sum\nolimits_{(\widehat{i}',\widehat{j}')\in\mathcal{M}_f}(\mathcal{E}(\widehat{i}', \widehat{j}',\mathbf{E})\cdot\mathbbm{1}(\sqrt{\mathcal{E}(\widehat{i}', \widehat{j}',\mathbf{E})})<\theta_f))\nonumber\\ &+ (\theta_f\cdot\mathbbm{1}(\sqrt{\mathcal{E}(\widehat{i}', \widehat{j}',\mathbf{E})})\geq\theta_f)),
\end{align}
where $\mathbf{E}$ represents the ground-truth essential matrix, $\mathbbm{1}(\cdot)$ is the Iverson bracket, $\theta_f=\frac{1.5}{(f_{x}^{\mathbf{A}}+f_{y}^{\mathbf{A}}+f_{x}^{\mathbf{B}}+f_{y}^{\mathbf{B}})}$ where $f_{x}^{\mathbf{I}}$ and $f_{y}^{\mathbf{I}}$ denote the focal lengths of image $\mathbf{I}$ corresponding to the $x$ and $y$ axes, respectively. Additionally,  $\mathcal{E}(\widehat{i}', \widehat{j}',\mathbf{E})$ is the Sampson distance~\cite{hartley2003multiple} which measures the geometric error of the match $(\widehat{i}',\widehat{j}')$ \emph{w.r.t}. $\mathbf{E}$, and its detailed definition is given in the \textbf{Supplementary Material} (\emph{\textbf{S.M.}}). The fine-level refinement loss is balanced as $\mathcal{L}_f=\alpha\mathcal{L}_{f1}+\beta\mathcal{L}_{f2}$.

\noindent{\textbf{Covisibility Estimation Supervision.}} To train the $\rm{MLP}$ of Eq.~\eqref{covisibility_prediction}, we first generate ground truth labels in light of two-view transformations. Specifically, a grid-level point in one image is covisible if, after warping with camera poses and depth maps, it falls within the other image and satisfies a depth consistency. We then minimize the binary cross-entropy of classifiers of layers $\ell\in\{2,\dots,L\}$, \emph{i.e.}, $\mathcal{L}_{covi}$.

Ultimately, we formulate the total loss as the weighted sum of all supervisions: $\mathcal{L}=\mathcal{L}_c+\mathcal{L}_f+\gamma\mathcal{L}_{covi}$.

\section{Experiments with CoMatch}

\subsection{Implementation Details}
We train CoMatch from scratch on a large-scale outdoor dataset MegaDepth~\cite{li2018megadepth}.  Training details are provided in the \emph{\textbf{S.M.}}. All images are resized to ensure their longer edge is 832 pixels. We leverage $L=4$ DCAT blocks to transform coarse-level features. For each CGTC module, we condense tokens with a kernel size of $4\times4$. $\tau$ is initialized to 10, while $\theta_c=0.1$. The loss function's weights are set to $\alpha=1.0$, $\beta=0.25$, and $\gamma=0.25$.  To showcase CoMatch's generalizability, we use the model trained on MegaDepth to evaluate all datasets and tasks in our experiments.

\subsection{Relative Pose Estimation}
\noindent{\textbf{Datasets.}} We leverage MegaDepth~\cite{li2018megadepth} and ScanNet~\cite{dai2017scannet} to demonstrate the effectiveness of CoMatch on camera pose estimation in outdoor and indoor scenes. Concretely, MegaDepth~\cite{li2018megadepth} comprises sparse 3D reconstructions of 196 popular landmarks, derived by COLMAP+MVS~\cite{schonberger2016structure,schonberger2016pixelwise}. The severe challenges of MegaDepth can be characterized by extreme viewpoint changes and repetitive patterns. Following~\cite{sun2021loftr,chen2022aspanformer}, we use the MegaDepth-1500 test set for evaluation. For semi-dense methods, we provide resized images with their longest dimension equal to 1152 pixels while for sparse ones, we resize images so that their longest dimension equals 1600 pixels. For dense matchers DKM and RoMa, we follow their evaluation setups and resize the input images to $880\times660$ and $672\times672$, respectively. ScanNet consists of 1613 monocular sequences with ground-truth camera poses and depth maps. It remains incredibly challenging due to wide baselines and texture scarcity. We evaluate our method on 1500 test pairs selected by~\cite{sarlin2020superglue}, where images are resized to $640\times480$ for all methods.

\noindent{\textbf{Baselines.}} For a well-rounded comparison, the baselines cover three categories: 1) sparse keypoint detection-description-matching methods, including SuperPoint (SP)~\cite{detone2018superpoint} coupled with the nearest neighbor (NN), IMP~\cite{xue2023imp}, SuperGlue (SG)~\cite{sarlin2020superglue}, LightGlue (LG)~\cite{lindenberger2023lightglue}, and DiffGlue~\cite{zhang2024diffglue}, 2) semi-dense matchers, including LoFTR~\cite{sun2021loftr}, Quadtree~\cite{tangquadtree}, MatchFormer~\cite{wang2022matchformer}, TopicFM~\cite{giang2023topicfm}, ASpanFormer~\cite{chen2022aspanformer}, and EfficientLoFTR~\cite{wang2024efficient}, and 3) dense matchers, including DKM~\cite{edstedt2023dkm} and RoMa~\cite{edstedt2024roma}.

\noindent{\textbf{Evaluation Protocols.}} In accordance with~\cite{sun2021loftr,chen2022aspanformer}, we calculate the area under the cumulative error curve (AUC) of the pose error (\emph{i.e.}, the maximum angular error in rotation and translation) at multiple thresholds (5$^\circ$, 10$^\circ$, and 20$^\circ$) to reflect the matching accuracy. Note that we retrieve relative camera motions by estimating an essential matrix via RANSAC and decomposing it into rotation and translation accordingly. Additionally, the processing time for matching each image pair in MegaDepth on a single NVIDIA RTX 4090 GPU is concluded to minutely understand the trade-off between matching accuracy and efficiency.

\begin{figure}[t]
	\centering
	\includegraphics[width=\linewidth]{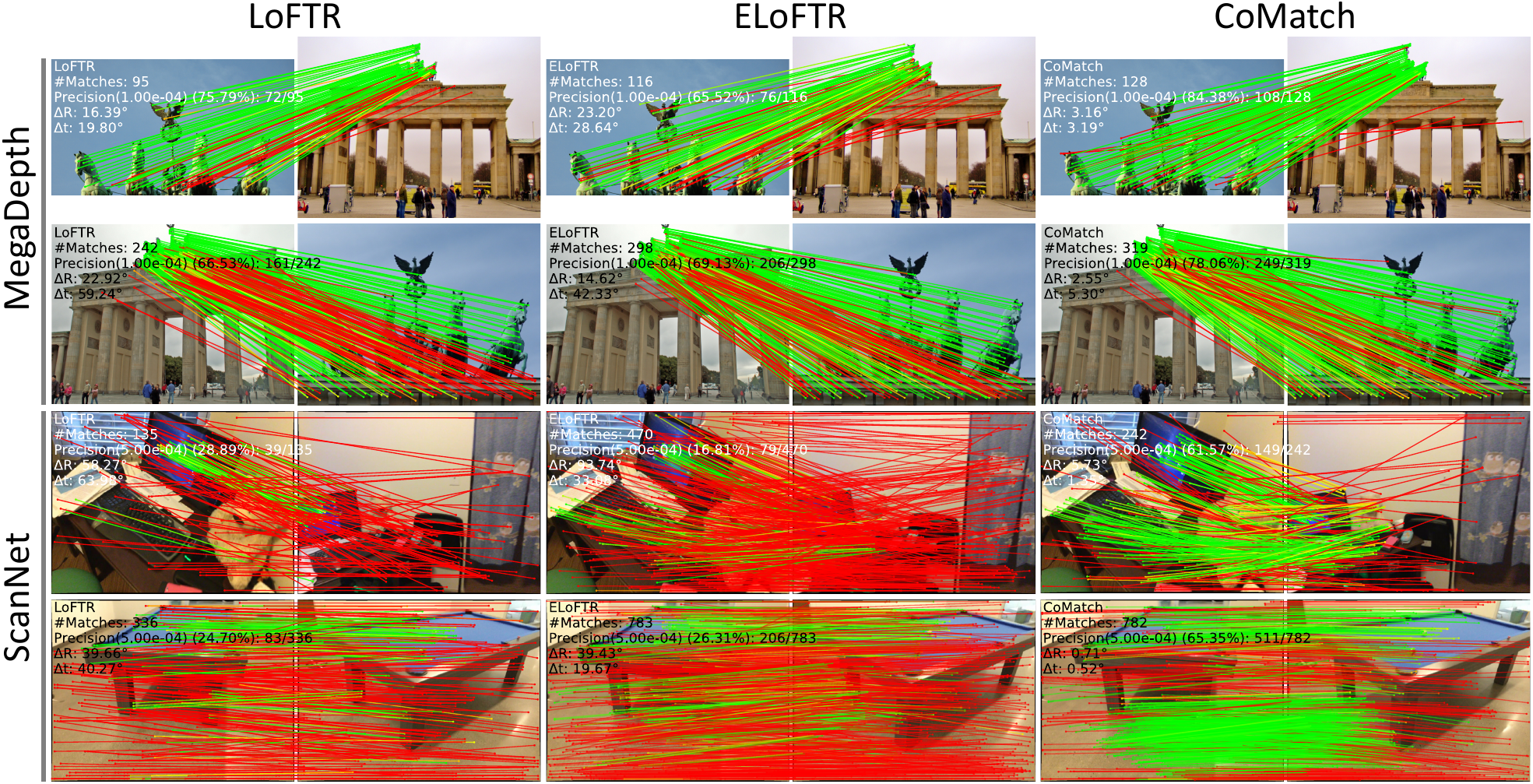}
	\caption{\textbf{Visualization of matching results on MegaDepth and ScanNet.} A match is ``\textcolor{green}{\raisebox{0.2\baselineskip}{\rule{0.5cm}{0.5mm}}}'' if its epipolar error is below $1\times10^{-4}$ for MegaDepth and $5\times10^{-4}$ for ScanNet, and ``\textcolor{red}{\raisebox{0.2\baselineskip}{\rule{0.5cm}{0.5mm}}}'' otherwise.}\label{matching_results}
\end{figure}

\noindent{\textbf{Results.}} The quantitative results are concluded in Tab.~\ref{relative_pose_estimation}, where CoMatch consistently outperforms existing sparse and semi-dense baselines on both datasets. Particularly, it largely excels the strong prior model ASpanFormer by 3.7\% in terms of AUC@5$^\circ$ while reducing the inference time by 41.5\% on MegaDepth. Despite being trained exclusively on outdoor scenes, our model shows promising accuracy on ScanNet, underscoring its excellent cross-dataset generalizability. Further looking at the visualization results illustrated in Fig.~\ref{matching_results}, CoMatch yields better correspondences and more accurate relative poses. Note that by drawing broad knowledge from the pre-trained vision foundation model DINOv2, RoMa is the most accurate among all evaluated methods but is slow for practical applications.  In contrast, CoMatch is over 6$\times$ faster with comparable performance, showcasing the best trade-off between accuracy and speed.

\begin{table}[t]
	\centering
    \renewcommand \arraystretch{1.2}
	\caption{\textbf{Evaluation on MegaDepth and ScanNet for relative pose estimation.} AUC of the pose error at different thresholds, along with the average runtime required to match a pair of images on MegaDepth, is reported.  The superscript $*$ denotes our re-implemented version. (\textcolor[RGB]{255, 199, 206}{\textbf{Red}}: optimal,  \textcolor[RGB]{217, 217, 255}{\textbf{Purple}}: suboptimal)}\label{relative_pose_estimation}
	\vspace{-0.05in}\resizebox{\linewidth}{!}{
		\begin{tabular}{ccccc}
			\toprule
			\multirow{2}{*}{\textbf{Category}}&\multirow{2}{*}{\textbf{Method}} & \multicolumn{1}{c}{\textbf{MegaDepth}}                & \textbf{ScanNet}  &     \multirow{2}{*}{\textbf{Time (ms)}\thinspace$\downarrow$}      \\ \cmidrule{3-4}
			&& \multicolumn{2}{c}{\textbf{AUC@5$^\circ$ / 10$^\circ$ / 20$^\circ$}\thinspace$\uparrow$} &   \\
			\midrule
			\multirow{5}{*}{{\textbf{Sparse}}}&SP + NN$_{\text{(CVPRW 18)}}$& \multicolumn{1}{c}{31.7 / 46.8 / 60.1} & 7.5 / 18.6 / 32.1& \colorbox[HTML]{FFC7CE}{28.5}\\
			&SP + IMP$_{\text{(CVPR 23)}}$& \multicolumn{1}{c}{44.9 / 62.5 / 76.4} & 15.2 / 31.8 / 48.4&102.5 \\
			&SP + SG$_{\text{(CVPR 20)}}$& \multicolumn{1}{c}{49.7 / \colorbox[HTML]{D9D9FF}{67.1} / \colorbox[HTML]{FFC7CE}{80.6}} & \colorbox[HTML]{FFC7CE}{16.2} / \colorbox[HTML]{FFC7CE}{32.8} / \colorbox[HTML]{FFC7CE}{49.7}&147.6 \\
			&SP + LG$_{\text{(ICCV 23)}}$& \multicolumn{1}{c}{\colorbox[HTML]{D9D9FF}{49.9} / 67.0 / \colorbox[HTML]{D9D9FF}{80.1}} & 14.8 / 30.8 / 47.5&\colorbox[HTML]{D9D9FF}{51.5}\\
			&SP + DG$_{\text{(ACM MM 24)}}$& \multicolumn{1}{c}{\colorbox[HTML]{FFC7CE}{50.2} / \colorbox[HTML]{FFC7CE}{67.3} / \colorbox[HTML]{D9D9FF}{80.1}} & \colorbox[HTML]{D9D9FF}{15.4} / \colorbox[HTML]{D9D9FF}{31.9} / \colorbox[HTML]{D9D9FF}{48.8}& 60.7 \\
			\hdashline
			\multirow{7}{*}{{\textbf{Semi-Dense}}}
			&LoFTR$_{\text{(CVPR 21)}}$&\multicolumn{1}{c}{52.8 / 69.2 / 81.2} &16.9 / 33.6 / 50.6&187.3 \\
			&QuadTree$_{\text{(ICLR 22)}}$&54.6 / 70.5 / 82.2&19.0 / 37.3 / 53.5&286.7 \\
			&MatchFormer$_{\text{(ACCV 22)}}$& \multicolumn{1}{c}{53.3 / 69.7 / 81.8} & 15.8 / 32.0 / 48.0&374.0\\
			&TopicFM$_{\text{(AAAI 23)}}$& \multicolumn{1}{c}{54.1 / 70.1 / 81.6} & 17.3 / 35.5 / 50.9&180.9\\
			&ASpanFormer$_{\text{(ECCV 22)}}$& \multicolumn{1}{c}{55.3 / 71.5 / 83.1} & \colorbox[HTML]{D9D9FF}{19.6} / \colorbox[HTML]{D9D9FF}{37.7} / \colorbox[HTML]{D9D9FF}{54.4}&211.8 \\
			&ELoFTR$_{\text{(CVPR 24)}}$& \multicolumn{1}{c}{\colorbox[HTML]{D9D9FF}{56.4} / \colorbox[HTML]{D9D9FF}{72.2} / \colorbox[HTML]{D9D9FF}{83.5}} & 19.2 / 37.0 / 53.6&\colorbox[HTML]{FFC7CE}{99.1}\\
			&ELoFTR*& \multicolumn{1}{c}{54.7 / 71.9 / 82.5} & 18.2 / 35.7 / 52.6 &-\\
			&CoMatch (ours)& \multicolumn{1}{c}{\colorbox[HTML]{FFC7CE}{58.0} / \colorbox[HTML]{FFC7CE}{73.2} / \colorbox[HTML]{FFC7CE}{84.2}} &\colorbox[HTML]{FFC7CE}{21.7} / \colorbox[HTML]{FFC7CE}{40.2} / \colorbox[HTML]{FFC7CE}{56.7}&\colorbox[HTML]{D9D9FF}{123.8} \\
			\hdashline
			\multirow{2}{*}{{\textbf{Dense}}}&DKM$_{\text{(CVPR 23)}}$& \multicolumn{1}{c}{60.4 / 74.9 / 85.1} & 26.6 / 47.1 / 64.2&588.9 \\
			&RoMa$_{\text{(CVPR 24)}}$& \multicolumn{1}{c}{62.6 / 76.7 / 86.3} &28.9 / 50.4 / 68.3&760.9 \\
			\bottomrule
	\end{tabular}}
\end{table}

\subsection{Homography Estimation}
\noindent{\textbf{Dataset.}} We adopt HPatches~\cite{balntas2017hpatches} to verify the efficacy of CoMatch on homography estimation, whcih comprises 52 sequences changing significantly in viewpoint and 56 sequences varying dramatically in illumination. Each sequence consists of 1 reference image
and 5 query ones, along with ground-truth homographies. All images are resized to ensure their shortest edge length of 480 pixels.

\noindent{\textbf{Baselines.}} We compare CoMatch comprehensively with two categories of methods: 1) detector-based sparse matchers, including IMP~\cite{xue2023imp}, SG~\cite{sarlin2020superglue}, LG~\cite{lindenberger2023lightglue}, and DG~\cite{zhang2024diffglue}, and 2) detector-free semi-dense matchers, including Sparse-NCNet~\cite{rocco2020efficient}, DRC-Net~\cite{li2020dual}, Quadtree~\cite{tangquadtree}, LoFTR~\cite{sun2021loftr}, ASpanFormer~\cite{chen2022aspanformer}, and EfficientLoFTR~\cite{wang2024efficient}. Note that all sparse matchers are paired with SP~\cite{detone2018superpoint} local features.

\noindent{\textbf{Evaluation Protocols.}} Following~\cite{detone2018superpoint}, we measure the mean reprojection error of 4 corner points between the images warped with the estimated homography $\hat{\mathcal{H}}$ and ground-truth homography $\mathcal{H}$ as a correctness identifier, and report AUC of the corner error up to thresholds of 3, 5, and 10 pixels to reflect the estimation accuracy. To ensure fairness, we extract up to 1024 local features for sparse matchers while merely retaining the top 1K predicted matches for semi-dense ones. Homographies are estimated via RANSAC.

\noindent{\textbf{Results.}} Tab.~\ref{HPatches} presents the all-sided quantitative evaluation on HPatches. Surprisingly, even though the match number is bound to 1K, CoMatch substantially advances sparse baselines across all error thresholds and sets state-of-the-art scores. This encouraging performance is attributed to the robust, compact, and efficient covisibility-dependent context aggregation enabled by the CGTC and CAA modules, alongside bilateral subpixel accuracy for both source and target views provided by the BSR module.

\begin{table}[t]
	\centering
    \renewcommand \arraystretch{1.15}
	\caption{\textbf{Evaluation on HPatches for homography estimation.} AUC of the corner error at different thresholds is reported.}
	\vspace{-0.05in}
	\label{HPatches}
	\resizebox{0.9\linewidth}{!}{
		\begin{tabular}{ccccc}
			\toprule
			\multirow{2}{*}{\textbf{Category}}&\multirow{2}{*}{\textbf{Method}}&\multicolumn{3}{c}{\textbf{HPathces}} \\
			\cmidrule{3-5}
			&&\textbf{AUC}@\textbf{3px}\thinspace$\uparrow$ &\textbf{AUC}@\textbf{5px}\thinspace$\uparrow$ &\textbf{AUC}@\textbf{10px}\thinspace$\uparrow$ \\
			\midrule
			\multirow{4}{*}{{\textbf{Sparse}}}
			&SP+IMP&56.1&68.2&80.9 \\
			&SP+SG&56.2&\colorbox[HTML]{D9D9FF}{68.5}&\colorbox[HTML]{D9D9FF}{81.1}\\
			&SP+LG&\colorbox[HTML]{D9D9FF}{56.4}&\colorbox[HTML]{D9D9FF}{68.5}&81.0 \\
			&SP+DG&\colorbox[HTML]{FFC7CE}{57.6}&\colorbox[HTML]{FFC7CE}{69.7}&\colorbox[HTML]{FFC7CE}{81.4} \\
			\hdashline
			\multirow{7}{*}{{\textbf{Semi-Dense}}}&Sparse-NCNet&48.9&54.2&67.1 \\
			&DRC-Net&50.6&56.2&68.3 \\
			&LoFTR&65.9&75.6&84.6 \\
			&Quadtree&66.3&76.2&84.9 \\
			&ASpanFormer&\colorbox[HTML]{D9D9FF}{67.4}&\colorbox[HTML]{D9D9FF}{76.9}&\colorbox[HTML]{D9D9FF}{85.6} \\
			&ELoFTR&66.5&76.4&85.5 \\
			&CoMatch (ours)&\colorbox[HTML]{FFC7CE}{68.4}&\colorbox[HTML]{FFC7CE}{78.2}&\colorbox[HTML]{FFC7CE}{86.8} \\
			\bottomrule
	\end{tabular}}
\end{table}

\subsection{Visual Localization}
\noindent{\textbf{Datasets.}} We adopt Aachen Day-Night v1.1~\cite{sattler2018benchmarking} and InLoc~\cite{taira2018inloc} datasets to evaluate the performance of CoMatch on visual localization. Specifically speaking, Aachen Day-Night v1.1 is a large-scale outdoor dataset featuring day-and-night illumination variations and large viewpoint changes. It includes 6697 reference images and 1015 (824 daytime, 191 nighttime) query images. We leverage its full localization
track for benchmarking. InLoc is a highly challenging indoor dataset characterized by repetitive patterns and poor textures, containing 9972 geometrically registered RGBD indoor images and 329 RGB query images annotated with manually verified camera poses. We select DUC1 and DUC2 for evaluation, as in~\cite{sun2021loftr,chen2022aspanformer}.

\noindent{\textbf{Baselines.}} We compare CoMatch with detector-based approaches, including SP~\cite{detone2018superpoint}+SG~\cite{sarlin2020superglue} and SP+LG~\cite{lindenberger2023lightglue}, as well as detector-free ones, such as LoFTR~\cite{sun2021loftr}, TopicFM~\cite{giang2023topicfm}, ASpanFormer~\cite{chen2022aspanformer}, and EfficientLoFTR~\cite{wang2024efficient}.

\noindent{\textbf{Evaluation Protocols.}} We adopt an open-source hierarchical
localization framework Hloc~\cite{sarlin2019coarse}, following~\cite{lindenberger2023lightglue,sun2021loftr}. We report the percentage of correctly localized queries at different distance and orientation thresholds, \emph{i.e.}, (0.25m, 2$^\circ$), (0.5m, 5$^\circ$), and (5.0m, 10$^\circ$), to indicate the accuracy.

\noindent{\textbf{Results.}} For the sake of fairness, we meticulously comply
with the pipeline and evaluation settings of the online visual localization benchmark\footnote{\url{https://www.visuallocalization.net/benchmark}}. Comprehensive quantitative evaluations on Aachen Day-Night v1.1 and InLoc are reported in Tab.~\ref{viusal_localization}. Clearly, CoMatch exhibits the overall best results, highlighting the strong generalizability of our method on real-world challenging visual localization scenarios.

\begin{table}[t]
	\centering
    \renewcommand \arraystretch{1.2}
	\caption{\textbf{Evaluation on Aachen Day-Night v1.1 and InLoc for visual localization.} The percentage of correctly localized queries at different thresholds is reported.}\label{viusal_localization}
	\vspace{-0.05in}
	\resizebox{\linewidth}{!}{%
		\begin{tabular}{ccccc}
			\toprule
			\multirow{2}{*}{\textbf{Method}}                                       & \textbf{Day}                          & \textbf{Night}   & \textbf{DUC1} & \textbf{DUC2}                   \\ \cmidrule{2-5}
			&  \multicolumn{4}{c}{\textbf{(0.25m, 2$^\circ$) / (0.5m, 5$^\circ$) / (5.0m, 10$^\circ$)}\thinspace$\uparrow$}                       \\ \midrule
			SP+SG              & \multicolumn{1}{c}{89.8 / \colorbox[HTML]{D9D9FF}{96.1} / \colorbox[HTML]{FFC7CE}{99.4}}            & 77.0 / 90.6 / \colorbox[HTML]{FFC7CE}{100.0}  & 47.0 / 69.2 / 79.8 &  53.4 / 77.1 / 80.9    \\
			SP+LG       & \colorbox[HTML]{FFC7CE}{90.2} / 96.0 / \colorbox[HTML]{FFC7CE}{99.4}            & 77.0 / 91.1 / \colorbox[HTML]{FFC7CE}{100.0}      &  49.0 / 68.2 / 79.3   & 55.0 / 74.8 / 79.4    \\
			LoFTR              & \multicolumn{1}{c}{88.7 / 95.6 / 99.0}            & \colorbox[HTML]{FFC7CE}{78.5} / 90.6 / 99.0  &   47.5 / 72.2 / 84.8 & 54.2 / 74.8 / 85.5   \\
			TopicFM & \colorbox[HTML]{FFC7CE}{90.2} / 95.9 / 98.9            & 77.5 / 91.1 / 99.5 & \colorbox[HTML]{D9D9FF}{52.0} / \colorbox[HTML]{D9D9FF}{74.7} / \colorbox[HTML]{FFC7CE}{87.4} & 53.4 / 74.8 / 83.2\\		
			ASpanFormer                                         & 89.4 / 95.6 / 99.0           & 77.5 / \colorbox[HTML]{FFC7CE}{91.6} / 99.5    & 51.5 / 73.7 / 86.4 &55.0 / 74.0 / 81.7   \\
			ELoFTR & 89.6 / \colorbox[HTML]{FFC7CE}{96.2} / 99.0          &  77.0 / 91.1 / 99.5     & \colorbox[HTML]{D9D9FF}{52.0} / \colorbox[HTML]{D9D9FF}{74.7} / \colorbox[HTML]{D9D9FF}{86.9} & \colorbox[HTML]{D9D9FF}{58.0} / \colorbox[HTML]{D9D9FF}{80.9} / \colorbox[HTML]{FFC7CE}{89.3}\\
			CoMatch (ours) &  89.4 / 95.8 / 99.0       & \colorbox[HTML]{FFC7CE}{78.5} / \colorbox[HTML]{FFC7CE}{91.6} / 99.5     &\colorbox[HTML]{FFC7CE}{54.5} / \colorbox[HTML]{FFC7CE}{75.3} / \colorbox[HTML]{D9D9FF}{86.9}	 & \colorbox[HTML]{FFC7CE}{59.5} / \colorbox[HTML]{FFC7CE}{84.7} / \colorbox[HTML]{D9D9FF}{87.8} \\
			\bottomrule
		\end{tabular}
	}
\end{table}

\subsection{Understanding CoMatch}\label{understanding_CoMatch}
\subsubsection{Ablation Studies}\label{ablation_studies}
In this part, we conduct holistic ablation studies to assess the validity of each design decision introduced in CoMatch. Results are summarized in Tab.~\ref{ablation_study}.

\noindent{\textbf{CGTC (row (a)).}} Replacing the CGTC module with a depth-wise convolution network and a max pooling layer for token condensing sparks a remarkable drop in AUC, suggesting that covisibility scores derived by Eq.~\eqref{covisibility_prediction} can provide explicit cues to guide token condensing, for generating ones with powerful representational capability.

\noindent{\textbf{CAA (rows (b)-(d)).}} Compared to vanilla / flash / linear attention, our CAA module effectively suppresses the interference of non-covisible regions, empowering CoMatch to achieve sharper and cleaner context interaction. In low-resolution scenes (ScanNet), the CAA module runs faster than linear attention.

\noindent{\textbf{BSR (rows (e)-(g)).}} To verify the positive effect of the second-stage refinement in our BSR module, we replace it with LoFTR's way or detaching it from CoMatch. Results reveal that the second stage contributes notably to the final performance, attributed to matches with bilateral subpixel accuracy favoring keypoint location-sensitive usages. We also completely remove the BSR module from CoMatch to highlight the necessity of match refinement.

\noindent{\textbf{Backbone (rows (h) and (i)).}} Replacing our backbone with RepVGG~\cite{ding2021repvgg} / ResNet-FPN~\cite{lin2017feature} just shows similar accuracy and efficiency, indicating that our design choice for local feature extraction is sound.

\begin{table}[!t]
	\centering
    \renewcommand \arraystretch{1.2}
	\caption{\textbf{Ablation studies.} For a comprehensive understanding of CoMatch, each component is ablated on ScanNet, where AUC@$5^{\circ}$/$10^{\circ}$/$20^{\circ}$ as well as the average runtime required to match an image pair at a resolution of $640\times480$ are reported.}
	\vspace{-0.05in}
	\label{ablation_study}
	\resizebox{\linewidth}{!}{
		\begin{tabular}{lcccc}
			\toprule
			\multirow{2}{*}{\textbf{Architecture}}&\multicolumn{3}{c}{\textbf{Pose estimation AUC}}&\multirow{2}{*}{\textbf{Time(ms)}\thinspace$\downarrow$} \\
			\cmidrule{2-4}
			&\textbf{@5$^{\circ}$}\thinspace$\uparrow$ &\textbf{@10$^{\circ}$}\thinspace$\uparrow$ &\textbf{@20$^{\circ}$}\thinspace$\uparrow$& \\
			\midrule
			(a) Without CGTC&20.0&37.7&54.6&33.4 \\
			\hdashline
			(b) Replace our CAA with vanilla attention&20.7&38.9&55.8&34.2 \\
			(c) Replace our CAA with flash attention&20.5&38.8&55.9&32.7\\
			(d) Replace our CAA with linear attention&18.1&34.8&51.0&35.3\\
			\hdashline
			(e) Replace 2nd-stage refine. with LoFTR’s refine.&20.4&38.7&55.2&33.3 \\
			(f) Without 2nd-stage refine.&20.2&37.9&54.9&31.9\\
			(g) Without the BSR module&11.3&24.1&37.4&\colorbox[HTML]{FFC7CE}{25.8} \\
			\hdashline
			(h) Replace our backbone with RepVGG&21.2&38.9&55.5&34.0 \\
			(i) Replace our backbone with ResNet-FPN&21.5&40.1&56.6&36.2 \\
			\midrule
			\textbf{CoMatch} (full)&\colorbox[HTML]{FFC7CE}{21.8}&\colorbox[HTML]{FFC7CE}{40.2}&\colorbox[HTML]{FFC7CE}{56.7}&34.5 \\
			\bottomrule
	\end{tabular}}
\end{table}

\begin{table}[!t]
	\centering
    \renewcommand \arraystretch{1.2}
	\caption{\textbf{Impact of test image resolution.} AUC of the pose error at multiple thresholds, along with the average runtime required to match an image pair at a specific resolution, is reported.}
	\vspace{-0.05in}
	\label{image_resolution}
	\resizebox{\linewidth}{!}{
		\begin{tabular}{ccccccccc}
			\toprule
			\multirow{2}{*}{{Resolution}}&\multicolumn{2}{c}{{ASpanFormer}}&&\multicolumn{2}{c}{{ELoFTR}}&&\multicolumn{2}{c}{{CoMatch}}\\
			\cmidrule{2-3} \cmidrule{5-6} \cmidrule{8-9}
			&{AUC@5$^{\circ}$ / 10$^{\circ}$ / 20$^{\circ}$}\thinspace$\uparrow$&{Time(ms)}\thinspace$\downarrow$&&{AUC@5$^{\circ}$ / 10$^{\circ}$ / 20$^{\circ}$}\thinspace$\uparrow$&{Time(ms)}\thinspace$\downarrow$&&{AUC@5$^{\circ}$ / 10$^{\circ}$ / 20$^{\circ}$}\thinspace$\uparrow$&{Time(ms)}\thinspace$\downarrow$ \\
			\midrule
			{640$\times$640}&\colorbox[HTML]{D9D9FF}{51.2} / \colorbox[HTML]{D9D9FF}{68.0} / \colorbox[HTML]{D9D9FF}{80.4}&70.3&&50.4 / 67.1 / 79.6&\colorbox[HTML]{FFC7CE}{31.4}&&\colorbox[HTML]{FFC7CE}{52.8} / \colorbox[HTML]{FFC7CE}{68.9} / \colorbox[HTML]{FFC7CE}{81.0}&\colorbox[HTML]{D9D9FF}{42.4} \\
			{800$\times$800}&53.2 / 69.7 / 81.8&114.6&&\colorbox[HTML]{D9D9FF}{53.6} / \colorbox[HTML]{D9D9FF}{70.2} / \colorbox[HTML]{D9D9FF}{82.1}&\colorbox[HTML]{FFC7CE}{44.7}&&\colorbox[HTML]{FFC7CE}{55.5} / \colorbox[HTML]{FFC7CE}{71.3} / \colorbox[HTML]{FFC7CE}{82.9}&\colorbox[HTML]{D9D9FF}{59.7} \\
			{960$\times$960}&54.6 / 70.8 / 82.5&143.1&&\colorbox[HTML]{D9D9FF}{54.9} / \colorbox[HTML]{D9D9FF}{71.2} / \colorbox[HTML]{D9D9FF}{82.8}&\colorbox[HTML]{FFC7CE}{65.0}&&\colorbox[HTML]{FFC7CE}{56.3} / \colorbox[HTML]{FFC7CE}{72.2} / \colorbox[HTML]{FFC7CE}{83.6}&\colorbox[HTML]{D9D9FF}{83.1} \\
			{1152$\times$1152}&55.3 / 71.5 / 83.1&211.8&&\colorbox[HTML]{D9D9FF}{56.4} / \colorbox[HTML]{D9D9FF}{72.2} / \colorbox[HTML]{D9D9FF}{83.5}&\colorbox[HTML]{FFC7CE}{99.1}&&\colorbox[HTML]{FFC7CE}{58.0} / \colorbox[HTML]{FFC7CE}{73.2} / \colorbox[HTML]{FFC7CE}{84.2}&\colorbox[HTML]{D9D9FF}{123.8} \\
			\bottomrule
	\end{tabular}}
\end{table}

\subsubsection{Impact of Test Image Resolution}\label{image_resolution_section}
In practice, the test image resolution is a critical factor that affects the  accuracy and efficiency of image matching. Consequently, we investigate the performance of our method at different image resolutions on MegaDepth, comparing with ASpanFormer~\cite{chen2022aspanformer} and ELoFTR~\cite{wang2024efficient}. Quantitative results are concluded in Tab.~\ref{image_resolution}. Notably, as image resolution increases, the accuracy of all methods improves, but at the cost of reduced efficiency. Evidently, CoMatch consistently achieves the best accuracy of pose estimation across all resolutions while maintaining the second-best efficiency. Surprisingly, CoMatch achieves superior performance at 800$\times$800 / 960$\times$960 resolution, outperforming advanced competitors operating at higher resolutions (960$\times$960 / 1152$\times$1152) with the faster speed. In a nutshell, CoMatch showcases commendable robustness to image resolution choices, making it well-suited for complicated and changeable real-life scenarios.

\section{Conclusion}
This paper presents CoMatch, a novel semi-dense matcher with superior accuracy, acceptable efficiency, and strong generalizability. Equipped with the CGTC and CAA modules, CoMatch enables message passing through a compact, robust, and efficient attention pattern. Also, the BSR module further endows CoMatch with the ability to produce geometrically consistent matches with bilateral subpixel accuracy. Extensive evaluations conducted across a variety of practical scenarios showcase CoMatch's superiority.

{
    \small
    \bibliographystyle{ieeenat_fullname}
    \bibliography{main}

\begin{thebibliography}{66}
\providecommand{\natexlab}[1]{#1}
\providecommand{\url}[1]{\texttt{#1}}
\expandafter\ifx\csname urlstyle\endcsname\relax
  \providecommand{\doi}[1]{doi: #1}\else
  \providecommand{\doi}{doi: \begingroup \urlstyle{rm}\Url}\fi

\bibitem[CVP()]{CVPR2022ImageMatching}
{CVPR} 2022 image matching challenge.
\newblock
  \url{https://www.kaggle.com/competitions/image-matching-challenge-2022/overview}.

\bibitem[Balntas et~al.(2017)Balntas, Lenc, Vedaldi, and
  Mikolajczyk]{balntas2017hpatches}
Vassileios Balntas, Karel Lenc, Andrea Vedaldi, and Krystian Mikolajczyk.
\newblock Hpatches: A benchmark and evaluation of handcrafted and learned local
  descriptors.
\newblock In \emph{CVPR}, pages 5173--5182, 2017.

\bibitem[Bay et~al.(2008)Bay, Ess, Tuytelaars, and Van~Gool]{bay2008speeded}
Herbert Bay, Andreas Ess, Tinne Tuytelaars, and Luc Van~Gool.
\newblock Speeded-up robust features (surf).
\newblock \emph{CVIU}, 110\penalty0 (3):\penalty0 346--359, 2008.

\bibitem[Cai et~al.(2022)Cai, Gan, and Han]{cai2022efficientvit}
Han Cai, Chuang Gan, and Song Han.
\newblock Efficientvit: Enhanced linear attention for high-resolution
  low-computation visual recognition.
\newblock \emph{arXiv preprint arXiv:2205.14756}, 3, 2022.

\bibitem[Cai et~al.(2024)Cai, Wang, Luo, Wang, Li, Xu, Gu, and
  Ai]{cai2024prism}
Xudong Cai, Yongcai Wang, Lun Luo, Minhang Wang, Deying Li, Jintao Xu, Weihao
  Gu, and Rui Ai.
\newblock Prism: Progressive dependency maximization for scale-invariant image
  matching.
\newblock In \emph{ACM MM}, 2024.

\bibitem[Carion et~al.(2020)Carion, Massa, Synnaeve, Usunier, Kirillov, and
  Zagoruyko]{carion2020end}
Nicolas Carion, Francisco Massa, Gabriel Synnaeve, Nicolas Usunier, Alexander
  Kirillov, and Sergey Zagoruyko.
\newblock End-to-end object detection with transformers.
\newblock In \emph{ECCV}, pages 213--229, 2020.

\bibitem[Chen et~al.(2021)Chen, Luo, Zhang, Zhou, Bai, Hu, Tai, and
  Quan]{chen2021learning}
Hongkai Chen, Zixin Luo, Jiahui Zhang, Lei Zhou, Xuyang Bai, Zeyu Hu, Chiew-Lan
  Tai, and Long Quan.
\newblock Learning to match features with seeded graph matching network.
\newblock In \emph{ICCV}, pages 6301--6310, 2021.

\bibitem[Chen et~al.(2022)Chen, Luo, Zhou, Tian, Zhen, Fang, Mckinnon, Tsin,
  and Quan]{chen2022aspanformer}
Hongkai Chen, Zixin Luo, Lei Zhou, Yurun Tian, Mingmin Zhen, Tian Fang, David
  Mckinnon, Yanghai Tsin, and Long Quan.
\newblock Aspanformer: Detector-free image matching with adaptive span
  transformer.
\newblock In \emph{ECCV}, pages 20--36, 2022.

\bibitem[Chen et~al.(2024{\natexlab{a}})Chen, Luo, Tian, Bai, Wang, Zhou, Zhen,
  Fang, Mckinnon, Tsin, et~al.]{chen2024affine}
Hongkai Chen, Zixin Luo, Yurun Tian, Xuyang Bai, Ziyu Wang, Lei Zhou, Mingmin
  Zhen, Tian Fang, David Mckinnon, Yanghai Tsin, et~al.
\newblock Affine-based deformable attention and selective fusion for semi-dense
  matching.
\newblock In \emph{CVPR}, pages 4254--4263, 2024{\natexlab{a}}.

\bibitem[Chen et~al.(2024{\natexlab{b}})Chen, Yu, Wan, Zhang, Wang, Zhong,
  Chen, and Yang]{chenecomatcher}
Peiqi Chen, Lei Yu, Yi Wan, Yongjun Zhang, Jian Wang, Liheng Zhong, Jingdong
  Chen, and Ming Yang.
\newblock Ecomatcher: Efficient clustering oriented matcher for detector-free
  image matching.
\newblock In \emph{ECCV}, 2024{\natexlab{b}}.

\bibitem[Dai et~al.(2017)Dai, Chang, Savva, Halber, Funkhouser, and
  Nie{\ss}ner]{dai2017scannet}
Angela Dai, Angel~X Chang, Manolis Savva, Maciej Halber, Thomas Funkhouser, and
  Matthias Nie{\ss}ner.
\newblock Scannet: Richly-annotated 3d reconstructions of indoor scenes.
\newblock In \emph{CVPR}, pages 5828--5839, 2017.

\bibitem[Deng et~al.(2024)Deng, Zhang, Zhang, Li, and Ma]{deng2024resmatch}
Yuxin Deng, Kaining Zhang, Shihua Zhang, Yansheng Li, and Jiayi Ma.
\newblock Resmatch: Residual attention learning for feature matching.
\newblock In \emph{AAAI}, pages 1501--1509, 2024.

\bibitem[DeTone et~al.(2018)DeTone, Malisiewicz, and
  Rabinovich]{detone2018superpoint}
Daniel DeTone, Tomasz Malisiewicz, and Andrew Rabinovich.
\newblock Superpoint: Self-supervised interest point detection and description.
\newblock In \emph{CVPRW}, pages 224--236, 2018.

\bibitem[Ding et~al.(2021)Ding, Zhang, Ma, Han, Ding, and Sun]{ding2021repvgg}
Xiaohan Ding, Xiangyu Zhang, Ningning Ma, Jungong Han, Guiguang Ding, and Jian
  Sun.
\newblock Repvgg: Making vgg-style convnets great again.
\newblock In \emph{CVPR}, pages 13733--13742, 2021.

\bibitem[Dosovitskiy(2020)]{dosovitskiy2020image}
Alexey Dosovitskiy.
\newblock An image is worth 16x16 words: Transformers for image recognition at
  scale.
\newblock In \emph{ICLR}, 2020.

\bibitem[Dusmanu et~al.(2019)Dusmanu, Rocco, Pajdla, Pollefeys, Sivic, Torii,
  and Sattler]{dusmanu2019d2}
Mihai Dusmanu, Ignacio Rocco, Tomas Pajdla, Marc Pollefeys, Josef Sivic,
  Akihiko Torii, and Torsten Sattler.
\newblock D2-net: A trainable cnn for joint description and detection of local
  features.
\newblock In \emph{CVPR}, pages 8092--8101, 2019.

\bibitem[Edstedt et~al.(2023)Edstedt, Athanasiadis, Wadenb{\"a}ck, and
  Felsberg]{edstedt2023dkm}
Johan Edstedt, Ioannis Athanasiadis, M{\aa}rten Wadenb{\"a}ck, and Michael
  Felsberg.
\newblock Dkm: Dense kernelized feature matching for geometry estimation.
\newblock In \emph{CVPR}, pages 17765--17775, 2023.

\bibitem[Edstedt et~al.(2024)Edstedt, Sun, B{\"o}kman, Wadenb{\"a}ck, and
  Felsberg]{edstedt2024roma}
Johan Edstedt, Qiyu Sun, Georg B{\"o}kman, M{\aa}rten Wadenb{\"a}ck, and
  Michael Felsberg.
\newblock Roma: Robust dense feature matching.
\newblock In \emph{CVPR}, pages 19790--19800, 2024.

\bibitem[Giang et~al.(2023)Giang, Song, and Jo]{giang2023topicfm}
Khang~Truong Giang, Soohwan Song, and Sungho Jo.
\newblock Topicfm: Robust and interpretable topic-assisted feature matching.
\newblock In \emph{AAAI}, pages 2447--2455, 2023.

\bibitem[Giang et~al.(2024)Giang, Song, and Jo]{giang2024topicfm+}
Khang~Truong Giang, Soohwan Song, and Sungho Jo.
\newblock Topicfm+: Boosting accuracy and efficiency of topic-assisted feature
  matching.
\newblock \emph{IEEE TIP}, 2024.

\bibitem[Hartley and Zisserman(2003)]{hartley2003multiple}
Richard Hartley and Andrew Zisserman.
\newblock \emph{Multiple view geometry in computer vision}.
\newblock Cambridge University Press, 2003.

\bibitem[He et~al.(2016)He, Zhang, Ren, and Sun]{he2016deep}
Kaiming He, Xiangyu Zhang, Shaoqing Ren, and Jian Sun.
\newblock Deep residual learning for image recognition.
\newblock In \emph{CVPR}, pages 770--778, 2016.

\bibitem[He et~al.(2024)He, Sun, Wang, Peng, Huang, Bao, and
  Zhou]{he2024detector}
Xingyi He, Jiaming Sun, Yifan Wang, Sida Peng, Qixing Huang, Hujun Bao, and
  Xiaowei Zhou.
\newblock Detector-free structure from motion.
\newblock In \emph{CVPR}, pages 21594--21603, 2024.

\bibitem[Huang et~al.(2023)Huang, Chen, Liu, Liu, Xu, Wu, Ding, Tang, and
  Wang]{huang2023adaptive}
Dihe Huang, Ying Chen, Yong Liu, Jianlin Liu, Shang Xu, Wenlong Wu, Yikang
  Ding, Fan Tang, and Chengjie Wang.
\newblock Adaptive assignment for geometry aware local feature matching.
\newblock In \emph{CVPR}, pages 5425--5434, 2023.

\bibitem[Jiang et~al.(2021)Jiang, Trulls, Hosang, Tagliasacchi, and
  Yi]{jiang2021cotr}
Wei Jiang, Eduard Trulls, Jan Hosang, Andrea Tagliasacchi, and Kwang~Moo Yi.
\newblock Cotr: Correspondence transformer for matching across images.
\newblock In \emph{ICCV}, pages 6207--6217, 2021.

\bibitem[Jiang et~al.(2023)Jiang, Zhang, Zhang, and Ma]{jiang2023improving}
Xingyu Jiang, Shihua Zhang, Xiao-Ping Zhang, and Jiayi Ma.
\newblock Improving sparse graph attention for feature matching by informative
  keypoints exploration.
\newblock \emph{CVIU}, 235:\penalty0 103803, 2023.

\bibitem[Katharopoulos et~al.(2020)Katharopoulos, Vyas, Pappas, and
  Fleuret]{katharopoulos2020transformers}
Angelos Katharopoulos, Apoorv Vyas, Nikolaos Pappas, and Fran{\c{c}}ois
  Fleuret.
\newblock Transformers are rnns: Fast autoregressive transformers with linear
  attention.
\newblock In \emph{ICML}, pages 5156--5165, 2020.

\bibitem[Kerbl et~al.(2023)Kerbl, Kopanas, Leimk{\"u}hler, and
  Drettakis]{kerbl20233d}
Bernhard Kerbl, Georgios Kopanas, Thomas Leimk{\"u}hler, and George Drettakis.
\newblock 3d gaussian splatting for real-time radiance field rendering.
\newblock \emph{ACM TOG}, 42\penalty0 (4):\penalty0 139--1, 2023.

\bibitem[Li et~al.(2020)Li, Han, Li, and Prisacariu]{li2020dual}
Xinghui Li, Kai Han, Shuda Li, and Victor Prisacariu.
\newblock Dual-resolution correspondence networks.
\newblock In \emph{NeurIPS}, pages 17346--17357, 2020.

\bibitem[Li et~al.(2021)Li, Si, Li, Hsieh, and Bengio]{li2021learnable}
Yang Li, Si Si, Gang Li, Cho-Jui Hsieh, and Samy Bengio.
\newblock Learnable fourier features for multi-dimensional spatial positional
  encoding.
\newblock In \emph{NeurIPS}, pages 15816--15829, 2021.

\bibitem[Li and Snavely(2018)]{li2018megadepth}
Zhengqi Li and Noah Snavely.
\newblock Megadepth: Learning single-view depth prediction from internet
  photos.
\newblock In \emph{CVPR}, pages 2041--2050, 2018.

\bibitem[Lin et~al.(2017)Lin, Doll{\'a}r, Girshick, He, Hariharan, and
  Belongie]{lin2017feature}
Tsung-Yi Lin, Piotr Doll{\'a}r, Ross Girshick, Kaiming He, Bharath Hariharan,
  and Serge Belongie.
\newblock Feature pyramid networks for object detection.
\newblock In \emph{CVPR}, pages 2117--2125, 2017.

\bibitem[Lindenberger et~al.(2023)Lindenberger, Sarlin, and
  Pollefeys]{lindenberger2023lightglue}
Philipp Lindenberger, Paul-Edouard Sarlin, and Marc Pollefeys.
\newblock Lightglue: Local feature matching at light speed.
\newblock In \emph{ICCV}, pages 17627--17638, 2023.

\bibitem[Lowe(2004)]{lowe2004distinctive}
David~G Lowe.
\newblock Distinctive image features from scale-invariant keypoints.
\newblock \emph{IJCV}, 60:\penalty0 91--110, 2004.

\bibitem[Luo et~al.(2020)Luo, Zhou, Bai, Chen, Zhang, Yao, Li, Fang, and
  Quan]{luo2020aslfeat}
Zixin Luo, Lei Zhou, Xuyang Bai, Hongkai Chen, Jiahui Zhang, Yao Yao, Shiwei
  Li, Tian Fang, and Long Quan.
\newblock Aslfeat: Learning local features of accurate shape and localization.
\newblock In \emph{CVPR}, pages 6589--6598, 2020.

\bibitem[Mildenhall et~al.(2021)Mildenhall, Srinivasan, Tancik, Barron,
  Ramamoorthi, and Ng]{mildenhall2021nerf}
Ben Mildenhall, Pratul~P Srinivasan, Matthew Tancik, Jonathan~T Barron, Ravi
  Ramamoorthi, and Ren Ng.
\newblock Nerf: Representing scenes as neural radiance fields for view
  synthesis.
\newblock \emph{CACM}, 65\penalty0 (1):\penalty0 99--106, 2021.

\bibitem[Mur-Artal and Tard{\'o}s(2017)]{mur2017orb}
Raul Mur-Artal and Juan~D Tard{\'o}s.
\newblock Orb-slam2: An open-source slam system for monocular, stereo, and
  rgb-d cameras.
\newblock \emph{IEEE TRO}, 33\penalty0 (5):\penalty0 1255--1262, 2017.

\bibitem[Mur-Artal et~al.(2015)Mur-Artal, Montiel, and Tardos]{mur2015orb}
Raul Mur-Artal, Jose Maria~Martinez Montiel, and Juan~D Tardos.
\newblock Orb-slam: a versatile and accurate monocular slam system.
\newblock \emph{IEEE TRO}, 31\penalty0 (5):\penalty0 1147--1163, 2015.

\bibitem[Ni et~al.(2023)Ni, Li, Huang, Li, Bao, Cui, and Zhang]{ni2023pats}
Junjie Ni, Yijin Li, Zhaoyang Huang, Hongsheng Li, Hujun Bao, Zhaopeng Cui, and
  Guofeng Zhang.
\newblock Pats: Patch area transportation with subdivision for local feature
  matching.
\newblock In \emph{CVPR}, pages 17776--17786, 2023.

\bibitem[Revaud et~al.(2019)Revaud, De~Souza, Humenberger, and
  Weinzaepfel]{revaud2019r2d2}
Jerome Revaud, Cesar De~Souza, Martin Humenberger, and Philippe Weinzaepfel.
\newblock R2d2: Reliable and repeatable detector and descriptor.
\newblock In \emph{NeurIPS}, 2019.

\bibitem[Rocco et~al.(2018)Rocco, Cimpoi, Arandjelovi{\'c}, Torii, Pajdla, and
  Sivic]{rocco2018neighbourhood}
Ignacio Rocco, Mircea Cimpoi, Relja Arandjelovi{\'c}, Akihiko Torii, Tomas
  Pajdla, and Josef Sivic.
\newblock Neighbourhood consensus networks.
\newblock In \emph{NeurIPS}, 2018.

\bibitem[Rocco et~al.(2020)Rocco, Arandjelovi{\'c}, and
  Sivic]{rocco2020efficient}
Ignacio Rocco, Relja Arandjelovi{\'c}, and Josef Sivic.
\newblock Efficient neighbourhood consensus networks via submanifold sparse
  convolutions.
\newblock In \emph{ECCV}, pages 605--621, 2020.

\bibitem[Rosten and Drummond(2006)]{rosten2006machine}
Edward Rosten and Tom Drummond.
\newblock Machine learning for high-speed corner detection.
\newblock In \emph{ECCV}, pages 430--443, 2006.

\bibitem[Rublee et~al.(2011)Rublee, Rabaud, Konolige, and
  Bradski]{rublee2011orb}
Ethan Rublee, Vincent Rabaud, Kurt Konolige, and Gary Bradski.
\newblock Orb: An efficient alternative to sift or surf.
\newblock In \emph{ICCV}, pages 2564--2571, 2011.

\bibitem[Sarlin et~al.(2019)Sarlin, Cadena, Siegwart, and
  Dymczyk]{sarlin2019coarse}
Paul-Edouard Sarlin, Cesar Cadena, Roland Siegwart, and Marcin Dymczyk.
\newblock From coarse to fine: Robust hierarchical localization at large scale.
\newblock In \emph{CVPR}, pages 12716--12725, 2019.

\bibitem[Sarlin et~al.(2020)Sarlin, DeTone, Malisiewicz, and
  Rabinovich]{sarlin2020superglue}
Paul-Edouard Sarlin, Daniel DeTone, Tomasz Malisiewicz, and Andrew Rabinovich.
\newblock Superglue: Learning feature matching with graph neural networks.
\newblock In \emph{CVPR}, pages 4938--4947, 2020.

\bibitem[Sattler et~al.(2018)Sattler, Maddern, Toft, Torii, Hammarstrand,
  Stenborg, Safari, Okutomi, Pollefeys, Sivic, et~al.]{sattler2018benchmarking}
Torsten Sattler, Will Maddern, Carl Toft, Akihiko Torii, Lars Hammarstrand,
  Erik Stenborg, Daniel Safari, Masatoshi Okutomi, Marc Pollefeys, Josef Sivic,
  et~al.
\newblock Benchmarking 6dof outdoor visual localization in changing conditions.
\newblock In \emph{CVPR}, pages 8601--8610, 2018.

\bibitem[Schonberger and Frahm(2016)]{schonberger2016structure}
Johannes~L Schonberger and Jan-Michael Frahm.
\newblock Structure-from-motion revisited.
\newblock In \emph{CVPR}, pages 4104--4113, 2016.

\bibitem[Sch{\"o}nberger et~al.(2016)Sch{\"o}nberger, Zheng, Frahm, and
  Pollefeys]{schonberger2016pixelwise}
Johannes~L Sch{\"o}nberger, Enliang Zheng, Jan-Michael Frahm, and Marc
  Pollefeys.
\newblock Pixelwise view selection for unstructured multi-view stereo.
\newblock In \emph{ECCV}, pages 501--518, 2016.

\bibitem[Shi et~al.(2022)Shi, Cai, Shavit, Mu, Feng, and
  Zhang]{shi2022clustergnn}
Yan Shi, Jun-Xiong Cai, Yoli Shavit, Tai-Jiang Mu, Wensen Feng, and Kai Zhang.
\newblock Clustergnn: Cluster-based coarse-to-fine graph neural network for
  efficient feature matching.
\newblock In \emph{CVPR}, pages 12517--12526, 2022.

\bibitem[Su et~al.(2024)Su, Ahmed, Lu, Pan, Bo, and Liu]{su2024roformer}
Jianlin Su, Murtadha Ahmed, Yu Lu, Shengfeng Pan, Wen Bo, and Yunfeng Liu.
\newblock Roformer: Enhanced transformer with rotary position embedding.
\newblock \emph{Neurocomputing}, 568:\penalty0 127063, 2024.

\bibitem[Sun et~al.(2021)Sun, Shen, Wang, Bao, and Zhou]{sun2021loftr}
Jiaming Sun, Zehong Shen, Yuang Wang, Hujun Bao, and Xiaowei Zhou.
\newblock Loftr: Detector-free local feature matching with transformers.
\newblock In \emph{CVPR}, pages 8922--8931, 2021.

\bibitem[Taira et~al.(2018)Taira, Okutomi, Sattler, Cimpoi, Pollefeys, Sivic,
  Pajdla, and Torii]{taira2018inloc}
Hajime Taira, Masatoshi Okutomi, Torsten Sattler, Mircea Cimpoi, Marc
  Pollefeys, Josef Sivic, Tomas Pajdla, and Akihiko Torii.
\newblock Inloc: Indoor visual localization with dense matching and view
  synthesis.
\newblock In \emph{CVPR}, pages 7199--7209, 2018.

\bibitem[Tang et~al.()Tang, Zhang, Zhu, and Tan]{tangquadtree}
Shitao Tang, Jiahui Zhang, Siyu Zhu, and Ping Tan.
\newblock Quadtree attention for vision transformers.
\newblock In \emph{ICLR}.

\bibitem[Truong et~al.(2020)Truong, Danelljan, and Timofte]{truong2020glu}
Prune Truong, Martin Danelljan, and Radu Timofte.
\newblock Glu-net: Global-local universal network for dense flow and
  correspondences.
\newblock In \emph{CVPR}, pages 6258--6268, 2020.

\bibitem[Truong et~al.(2021)Truong, Danelljan, Van~Gool, and
  Timofte]{truong2021learning}
Prune Truong, Martin Danelljan, Luc Van~Gool, and Radu Timofte.
\newblock Learning accurate dense correspondences and when to trust them.
\newblock In \emph{CVPR}, pages 5714--5724, 2021.

\bibitem[Truong et~al.(2023)Truong, Danelljan, Timofte, and
  Van~Gool]{truong2023pdc}
Prune Truong, Martin Danelljan, Radu Timofte, and Luc Van~Gool.
\newblock Pdc-net+: Enhanced probabilistic dense correspondence network.
\newblock \emph{IEEE TPAMI}, 45\penalty0 (8):\penalty0 10247--10266, 2023.

\bibitem[Vaswani(2017)]{vaswani2017attention}
A Vaswani.
\newblock Attention is all you need.
\newblock In \emph{NeurIPS}, 2017.

\bibitem[Wang et~al.(2023)Wang, Xu, Lu, Xu, Meng, Zhang, Fan, and
  Zhang]{wang2023attention}
Changwei Wang, Rongtao Xu, Ke Lu, Shibiao Xu, Weiliang Meng, Yuyang Zhang, Bin
  Fan, and Xiaopeng Zhang.
\newblock Attention weighted local descriptors.
\newblock \emph{IEEE TPAMI}, 45\penalty0 (9):\penalty0 10632--10649, 2023.

\bibitem[Wang et~al.(2022)Wang, Zhang, Yang, Peng, and
  Stiefelhagen]{wang2022matchformer}
Qing Wang, Jiaming Zhang, Kailun Yang, Kunyu Peng, and Rainer Stiefelhagen.
\newblock Matchformer: Interleaving attention in transformers for feature
  matching.
\newblock In \emph{ACCV}, pages 2746--2762, 2022.

\bibitem[Wang et~al.(2024)Wang, He, Peng, Tan, and Zhou]{wang2024efficient}
Yifan Wang, Xingyi He, Sida Peng, Dongli Tan, and Xiaowei Zhou.
\newblock Efficient loftr: Semi-dense local feature matching with sparse-like
  speed.
\newblock In \emph{CVPR}, pages 21666--21675, 2024.

\bibitem[Wu et~al.(2020)Wu, Pan, Chen, Long, Zhang, and
  Philip]{wu2020comprehensive}
Zonghan Wu, Shirui Pan, Fengwen Chen, Guodong Long, Chengqi Zhang, and S~Yu
  Philip.
\newblock A comprehensive survey on graph neural networks.
\newblock \emph{IEEE TNNLS}, 32\penalty0 (1):\penalty0 4--24, 2020.

\bibitem[Xue et~al.(2023{\natexlab{a}})Xue, Budvytis, and Cipolla]{xue2023imp}
Fei Xue, Ignas Budvytis, and Roberto Cipolla.
\newblock Imp: Iterative matching and pose estimation with adaptive pooling.
\newblock In \emph{CVPR}, pages 21317--21326, 2023{\natexlab{a}}.

\bibitem[Xue et~al.(2023{\natexlab{b}})Xue, Budvytis, and Cipolla]{xue2023sfd2}
Fei Xue, Ignas Budvytis, and Roberto Cipolla.
\newblock Sfd2: Semantic-guided feature detection and description.
\newblock In \emph{CVPR}, pages 5206--5216, 2023{\natexlab{b}}.

\bibitem[Zhang and Ma(2024)]{zhang2024diffglue}
Shihua Zhang and Jiayi Ma.
\newblock Diffglue: Diffusion-aided image feature matching.
\newblock In \emph{ACM MM}, 2024.

\bibitem[Zhao et~al.(2023)Zhao, Wu, Chen, Chen, Xu, and Li]{zhao2023aliked}
Xiaoming Zhao, Xingming Wu, Weihai Chen, Peter~CY Chen, Qingsong Xu, and
  Zhengguo Li.
\newblock Aliked: A lighter keypoint and descriptor extraction network via
  deformable transformation.
\newblock \emph{IEEE TIM}, 72:\penalty0 1--16, 2023.

\end{thebibliography}
}

\maketitlesupplementary
\renewcommand\thesection{\Alph{section}}
\setcounter{section}{0} 
\section{Implementation Details}
\subsection{Training Details}
We follow the same training-test split as LoFTR~\cite{sun2021loftr}. The network is trained for 30 epochs using the AdamW optimizer with an initial learning rate of $1\times10^{-3}$ and a batch size of 8. Training is conducted on 4 NVIDIA RTX 4090 GPUs and completes in approximately 22 hours.

\subsection{Architecture}
\subsubsection{Local Feature Extraction}
The local feature extraction is fleshed out by a modified ResNet-18~\cite{he2016deep} without the bottom-up part. Specifically, we use a width of 64 and a stride of 1 for the stem and widths of [64, 128, 256] and strides of 2 for the subsequent three stages. The output of the last stage at $\nicefrac{1}{8}$ image resolution is processed by our dynamic covisibility-aware Transformer (DCAT) to derive discriminative coarse-level features. The second and third stages' feature maps are at $\nicefrac{1}{2}$ and $\nicefrac{1}{4}$ image resolutions, respectively, which are progressively fused with transformed coarse-level features to produce cross-view perceived fine-level ones for subsequent match refinement. 

\subsubsection{Position Encoding}
The spatial location context is essential for matching, typically modeled by absolute positional encoding (PE)~\cite{carion2020end}. However, in projective camera geometry, the position of visual observations showcases equivariance concerning the camera's translation motion within the image plane~\cite{lindenberger2023lightglue}. This reveals that an encoding should exclusively consider the relative but not the absolute position of keypoints. To this end, we adopt Rotary position encoding (RoPE)~\cite{su2024roformer} to encode the spatial positional context between coarse-level features. More concretely, for each coarse feature $i$, we first decompose it into query and key vectors $\mathbf{q}_i$ and $\mathbf{k}_i$ via \emph{different} linear transformations, then the attention score between two coarse features $i$ and $j$ is defined as follows:
\begin{equation}\label{attention_score}
	a_{ij}=\mathbf{q}_i^{\top}\mathbf{R}(\mathbf{x}_j-\mathbf{x}_i)\mathbf{k}_j,
\end{equation}
where $\mathbf{x}_i$ and $\mathbf{x}_j$ are the 2D image coordinates of $\mathbf{q}_i$ and $\mathbf{k}_j$, respectively, and $\mathbf{R}(\cdot)\in\mathbbm{R}^{d\times d}$ is a block diagonal matrix encoding the relative position between coarse features. We partition the space into $d/2$ 2D subspaces and rotate each of them with an angle corresponding to the projection onto a learnable basis $\mathbf{b}_k\in\mathbbm{R}^2$, following Fourier Features~\cite{li2021learnable}:
\begin{equation}\label{rope}
	\mathbf{R}(\mathbf{x}) = 
	\begin{pmatrix}
		\hat{\mathbf{R}}(\mathbf{b}_1^\top \mathbf{x}) & \mathbf{0} & \cdots & \mathbf{0} \\
		\mathbf{0} & \hat{\mathbf{R}}(\mathbf{b}_2^\top \mathbf{x}) & \cdots & \mathbf{0} \\
		\vdots & \vdots & \ddots & \vdots \\
		\mathbf{0} & \mathbf{0} & \cdots & \hat{\mathbf{R}}(\mathbf{b}_{d/2}^\top \mathbf{x})
	\end{pmatrix}
\end{equation}
where $
\hat{\mathbf{R}}(\theta) = 
\begin{pmatrix}
	\cos \theta & -\sin \theta \\
	\sin \theta & \cos \theta
\end{pmatrix}$. 

By doing so, the model can retrieve coarse feature $j$ located at a learned relative position from $i$, concentrating more on interaction between features instead of their specific locations. Note that the encoding remains identical across all self-attention layers, allowing it to be computed once and then cached for reuse.

\subsection{Supervision}
\noindent{\textbf{Fine-Level Refinement Supervision.}} We supervise the second stage with an epipolar geometry loss $\mathcal{L}_{f2}$, as defined in Eq.~(8) of the main paper, where $\mathcal{E}(\widehat{i}', \widehat{j}',\mathbf{E})$ is the Sampson distance~\cite{hartley2003multiple} that measures the geometric error of the match $(\widehat{i}',\widehat{j}')$ \emph{w.r.t}. $\mathbf{E}$ and is defined as:
\begin{align}\label{fine_loss}
	&\mathcal{E}(\widehat{i}', \widehat{j}',\mathbf{E})\nonumber\\
	&=\frac{({\mathbf{p}_{\widehat{j}'}^{\top}}\mathbf{E}\mathbf{p}_{\widehat{i}'})^2}{\Vert\mathbf{E}\mathbf{p}_{\widehat{i}'}\Vert^2_{[1]} +\Vert\mathbf{E}\mathbf{p}_{\widehat{i}'}\Vert^2_{[2]}+\Vert\mathbf{E}^{\top}\mathbf{p}_{\widehat{j}'}\Vert^2_{[1]}+\Vert\mathbf{E}^{\top}\mathbf{p}_{\widehat{j}'}\Vert^2_{[2]}}.
\end{align}
Importantly, $\mathbf{p}_{\widehat{i}'}$ and $\mathbf{p}_{\widehat{j}'}$ are homogeneous coordinates of two keypoints $\widehat{i}'$ and $\widehat{j}'$ which form a final fine-level correspondence, and $\mathbf{v}_{[k]}$ denotes the $k$-th element of vector $\mathbf{v}$.

\begin{table}[!t]
	\centering
	\caption{\textbf{Image matching challenge.} mAA@10$^\circ$ of the pose error is reported. The superscript $*$ denotes our re-implemented version.}
	\vspace{-0.05in}
	\label{imc}
	\resizebox{0.6\columnwidth}{!}{%
		\begin{tabular}{l c}
			\toprule
			\textbf{Method} & \textbf{mAA}@10$^\circ$ $\uparrow$ \\
			\midrule
			LoFTR & 78.3 \\
			MatchFormer & 78.3 \\
			QuadTree & 81.2 \\
			ASpanFormer & 82.2 \\
			ELoFTR & 81.3 \\
			\textbf{CoMatch} (ours) & 82.3 \\
			\hdashline
			DKM & 82.6*/ 83.1 \\
			RoMa & \textbf{85.5*} / \textbf{88.0} \\
			\bottomrule
		\end{tabular}
	}
\end{table}

\begin{table*}[!t]
	\centering
	\caption{\textbf{Comparison with TopicFM+ and PATS.} The runtime to match an image pair on MegaDepth is reported. }\label{relative_pose_estimation}
	\vspace{-0.05in}
	\label{SOTA_comparison}
	\resizebox{0.8\linewidth}{!}{
		\begin{tabular}{ccccccc}
			\toprule
			\multirow{2}{*}{\textbf{Method}} & \multicolumn{1}{c}{\textbf{MegaDepth}}                & \textbf{ScanNet}  && \textbf{DUC1}                & \textbf{DUC2}  &     \multirow{2}{*}{\textbf{Time (ms)}\thinspace$\downarrow$}      \\ \cmidrule{2-3} \cmidrule{5-6}
			& \multicolumn{2}{c}{\textbf{AUC@5$^\circ$ / 10$^\circ$ / 20$^\circ$}\thinspace$\uparrow$} && \multicolumn{2}{c}{\textbf{(0.25m, 2$^\circ$) / (0.5m, 5$^\circ$) / (5.0m, 10$^\circ$)}\thinspace$\uparrow$}&  \\
			\midrule
			TopicFM+&54.2 / 70.5 / 82.5&20.4 / 38.3 / 54.6&&52.0 / 74.7 / \colorbox[HTML]{FFC7CE}{87.4} &53.4 / 74.8 / 83.2&135.6\\
			PATS&\colorbox[HTML]{FFC7CE}{61.0} / \colorbox[HTML]{FFC7CE}{74.2} / 83.0&20.9 / 40.1 / \colorbox[HTML]{FFC7CE}{57.2}&&\colorbox[HTML]{FFC7CE}{55.6} / 71.2 / 81.0 &58.8 / 80.9 / 85.5& 773.4\\
			CoMatch& 58.0 / 73.2 / \colorbox[HTML]{FFC7CE}{84.2} &\colorbox[HTML]{FFC7CE}{21.7} / \colorbox[HTML]{FFC7CE}{40.2} / 56.7&&54.5 / \colorbox[HTML]{FFC7CE}{75.3} / 86.9	 & \colorbox[HTML]{FFC7CE}{59.5} / \colorbox[HTML]{FFC7CE}{84.7} / \colorbox[HTML]{FFC7CE}{87.8}&\colorbox[HTML]{FFC7CE}{123.8} \\
			\bottomrule
	\end{tabular}}
\end{table*}

\begin{table}[!t]
	\centering
	\caption{\textbf{Quantitative evaluation of covisibility scores.}}
	\vspace{-0.05in}
	\label{covis}
	\resizebox{0.65\columnwidth}{!}{%
		\begin{tabular}{ccc}
			\toprule
			\textbf{Metric} & \textbf{Source View} & \textbf{Target View} \\
			\midrule
			Precision & 88.7 & 88.3 \\
			Recall & 83.8 & 84.3 \\
			\bottomrule
	\end{tabular}}
\end{table}

\section{Experiments}
\subsection{Image Matching Challenge}
To further substantiate CoMatch’s performance on relative pose estimation, we evaluate it on Kaggle competition Image Matching Challenge (IMC) 2022 benchmark~\cite{CVPR2022ImageMatching}, where the we estimate a fundamental matrix via RANSAC and decomposing it into rotation and translation accordingly.

\noindent{\textbf{Dataset.}} IMC 2022 offers a test set comprising roughly 10,000 Google Street View images that exhibit significant visual diversity. These images are captured from a wide range of viewpoints, featuring varied aspect ratios, lighting and weather conditions, and occlusions from both pedestrians and vehicles. Notably, the evaluation dataset remains undisclosed to participants, being securely hosted on Kaggle's competition platform to ensure fair benchmarking.

\noindent{\textbf{Baselines.}} We compare CoMatch with LoFTR~\cite{sun2021loftr}, MatchFormer~\cite{wang2022matchformer}, QuadTree~\cite{tangquadtree}, ASpanFormer~\cite{chen2022aspanformer}, ELoFTR~\cite{wang2024efficient}, DKM~\cite{edstedt2023dkm}, and RoMa~\cite{edstedt2024roma}.

\noindent{\textbf{Evaluation Protocol.}} We calculate the mean average accuracy (mAA) between the estimated fundamental matrix and the hidden ground-truth counterpart.  This assessment considers pose errors through two criteria: rotation deviation in degrees and translation discrepancy in meters. A pose is classified as accurate if it meets both thresholds. In IMC 2022, ten pairs of thresholds are considered: the rotation threshold ranges from 1$^\circ$ to 10$^\circ$ while the translation threshold spans 0.2 m to 5 m, with both thresholds following uniform distributions across their respective ranges. After that, the percentage of image pairs that meet every pair of thresholds can be determined, and the average of results over all
threshold pairs is mAA.

\noindent{\textbf{Results.}} Quantitative results on IMC 2022 are reported in Tab.~\ref{imc}, where CoMatch outperforms all semi-dense matchers. Compared to DKM and RoMa, CoMatch is much faster (see Tab.~1 of the main paper) with comparable performance, showing a better trade-off between accuracy and efficiency.

\subsection{Comparison with More Recent Baselines}
To further highlight the superiority of our CoMatch, we compare it with TopicFM+~\cite{giang2024topicfm+} and PATS~\cite{ni2023pats} on MegaDepth~\cite{li2018megadepth}, ScanNet~\cite{dai2017scannet}, and Inloc~\cite{taira2018inloc}. Results are presented in Tab.~\ref{SOTA_comparison}, where TopicFM+’s results on MegaDepth differs from the original since we adjust its 0.2 RANSAC threshold to standard 0.5 for fair comparison. Results show CoMa’s strong generalizability on ScanNet and Inloc. On MegaDepth, CoMatch outperforms PATS by being over 6$\times$ faster with comparable accuracy.

\subsection{Covisibility Quantitative Evaluation}
We predict soft covisibility scores per token via Eq. (2) instead of using hard masks to guide feature matching. To quantitatively evaluate the classifier, we adopt a threshold of 0.5 to classify tokens as covisible or non-covisible and compute its precision and recall on MegaDepth at 832$\times$832 resolution. As reported in Tab.~\ref{covis}, the classifier achieves (88.7, 83.8) and (88.3, 84.3) for two views, respectively, showing its reliability in guiding our CGTC and CAA modules toward robust and compact context interaction.

\begin{table}[!t]
	\centering
	\caption{\textbf{The average runtime per process required when matching an image pair on MegaDepth at a resolution of 1152$\times$1152.}}
	\vspace{-0.05in}
	\label{timing}
	\resizebox{\linewidth}{!}{
		\begin{tabular}{lc}
			\toprule
			{Process}&{{Time (ms)}\thinspace$\downarrow$} \\
			\midrule
			{(a) Local Feature Extraction}&9.7 \\
			{(b) Dynamic CovisibilityAware
				Transformer}&19.4 \\
			{(c) Coarse-Level
				Match Determination}&62.0 \\
			{(d) Bilateral Subpixel-Level
				Refinement}&32.7 (12.6 / 20.1)\\
			\midrule
			\textbf{Total}&123.8 \\
			\bottomrule
	\end{tabular}}
\end{table}

\subsection{Timing}
In the main paper, we average the runtime across all image pairs in the test dataset, \emph{i.e.}, MegaDepth~\cite{li2018megadepth}, for efficiency evaluation, with a warm-up of 50 pairs to ensure accurate measurement. All comparative methods are implemented on a single NVIDIA GeForce RTX 4090 with 32 cores of Intel(R) Xeon(R) Platinum 8336C CPU.

In this supplementary material, we further present the average runtime per procedure of CoMatch in Tab.~\ref{timing} for a more detailed efficiency analysis. We notice that a large fraction of time is spent on the coarse-level match determination, where a dual-softmax operation is used to generate the assignment matrix but may substantially increase the latency during the inference phase, particularly for high-resolution cases (\emph{i.e.}, the large number of tokens). As the bilateral subpixel-level refinement module comprises a progressive feature fusion layer and a two-stage correlation layer, we also report their average runtime in row (d) of Tab.~\ref{timing}.

\begin{table}[!t]
	\centering
	\caption{\textbf{Impact of condensing range on MegaDepth.} AUC of the pose error at multiple thresholds, together with the average runtime required to match an image pair at a resolution of 1152$\times$1152, is reported. The best results are in \textbf{bold}.}
	\vspace{-0.05in}
	\label{kernel_size}
	\resizebox{\linewidth}{!}{
		\begin{tabular}{ccc}
			\toprule
			\multirow{2}{*}{{Condensing Range}}&{{Pose Estimation AUC}}&\multirow{2}{*}{{Time (ms)}\thinspace$\downarrow$}\\
			\cmidrule{2-2}
			&{AUC@5$^{\circ}$ / 10$^{\circ}$ / 20$^{\circ}$}\thinspace$\uparrow$&\\
			\midrule
			{$s=2$}&57.3 / 73.0 / \textbf{84.2}&207.6 \\
			{$s=4$}&\textbf{58.0} / \textbf{73.2} / \textbf{84.2}&\textbf{123.8} \\
			\bottomrule
	\end{tabular}}
\end{table}

\subsection{Impact of Condensing Range}
Adaptively condensing tokens in light of their covisibility scores that are dynamically estimated within the network lays the foundation for the subsequent covisibility-assisted attention module. Thereby, we investigate the impact of different condensing ranges on the matching performance of CoMatch, with results presented in Tab.~\ref{kernel_size}, where $s=4$ serves as the default setting. Notably, employing a smaller condensing range, \emph{i.e.}, 2$\times$2, increases the number of reduced tokens, resulting in a slight drop in accuracy but significantly slower speed. This also underscores the rationality of our chosen condensing range parameter.

\begin{figure*}[t]
	\centering
	\includegraphics[width=\linewidth]{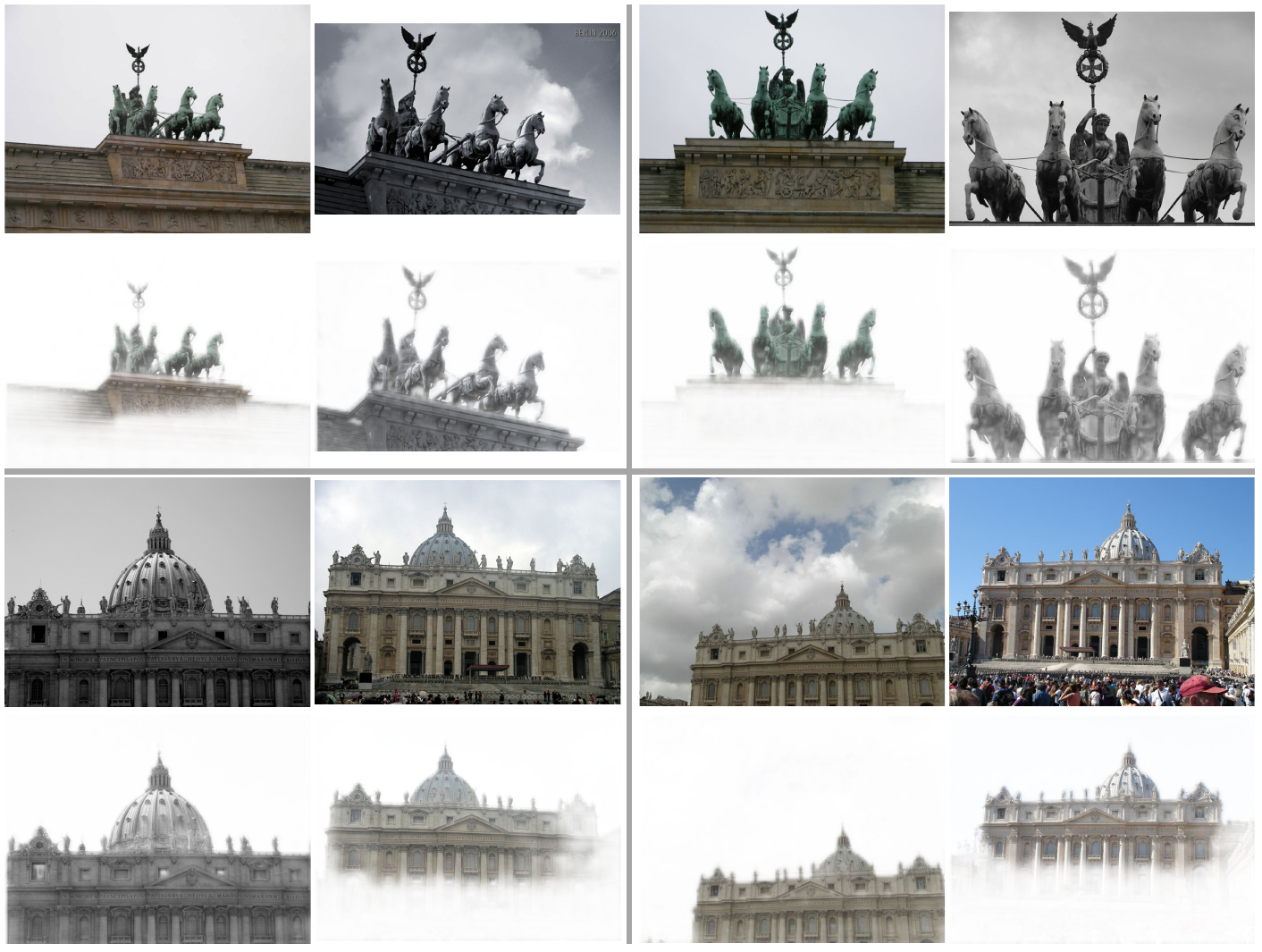}
	\vspace{-0.05in}
	\caption{\textbf{Visualization of covisibility prediction.} We first bilinearly
		up-sample the covisibility score map to match the original
		image resolution, and then multiply it with the input image.}\label{covis_sm}
\end{figure*}

\begin{figure*}[t]
	\centering
	\includegraphics[width=\linewidth]{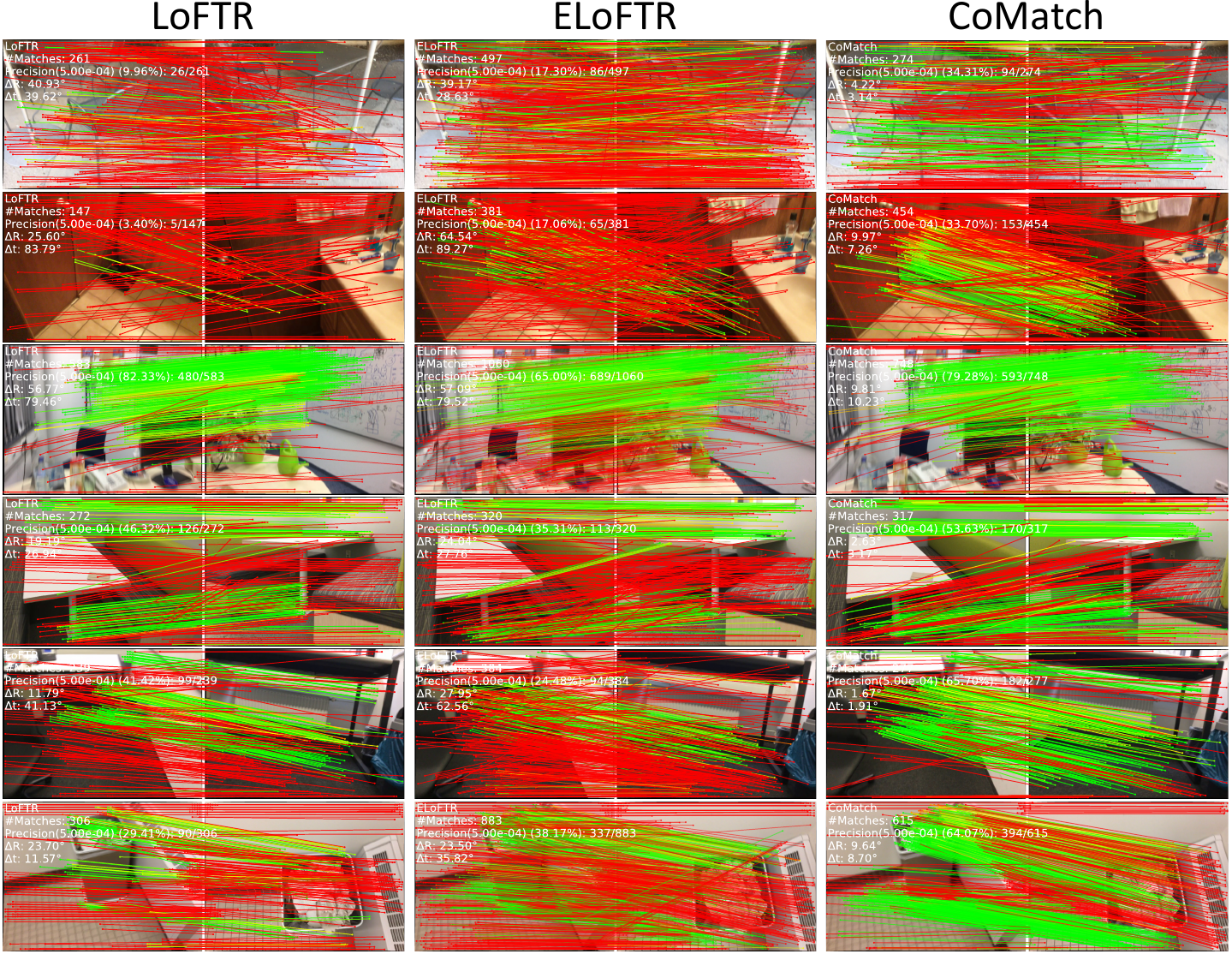}
	\vspace{-0.05in}
	\caption{\textbf{Visualization of matching results on ScanNet.} A match is ``\textcolor{green}{\raisebox{0.2\baselineskip}{\rule{0.5cm}{0.5mm}}}'' if its epipolar error is below $5\times10^{-4}$, and ``\textcolor{red}{\raisebox{0.2\baselineskip}{\rule{0.5cm}{0.5mm}}}'' otherwise.}\label{matching_sm}
\end{figure*}

\subsection{More Qualitative Results}
Fig.~\ref{covis_sm} illustrates the covisibility prediction of CoMatch on four representative examples. Evidently, our approach demonstrates the capability to precisely predict the covisible regions between image pairs, benefiting our covisibility-guided token condensing and covisibility-assisted attention. Fig.~\ref{matching_sm} presents the matching results on ScanNet~\cite{dai2017scannet}. Compared to LoFTR~\cite{sun2021loftr} and ELoFTR~\cite{wang2024efficient}, our CoMatch establishes more reliable correspondences and recovers more accurate relative camera poses.

\end{document}